%% file: main.tex
\documentclass[10pt,twocolumn,letterpaper]{article}

\usepackage[pagenumbers]{cvpr} 

\input{inc/inc_packages}

\input{inc/inc_macros}
\input{inc/math_commands}

\definecolor{cvprblue}{rgb}{0.21,0.49,0.74}
\usepackage[pagebackref,breaklinks,colorlinks,allcolors=cvprblue]{hyperref}
\usepackage{threeparttable}

\def\paperID{5448} 
\def\confName{CVPR}
\def\confYear{2025}

\title{T*: Re-thinking Temporal Search for Long-Form Video Understanding}

\author{
Jinhui Ye$^{1}$\thanks{Equal contribution.}~\thanks{Work done during internship at Stanford.}~, 
Zihan Wang$^{2*}$, Haosen Sun$^{2}$, Keshigeyan Chandrasegaran$^{1}$, \\
Zane Durante$^{1}$, Cristobal Eyzaguirre$^{1}$, Yonatan Bisk$^{3}$, Juan Carlos Niebles$^{1}$, 
Ehsan Adeli$^{1}$, \\
Li Fei-Fei$^{1}$, Jiajun Wu$^{1}$, Manling Li$^{1,2}$ \\[0.5em]
$^{1}$Stanford University\quad
$^{2}$Northwestern University\quad
$^{3}$Carnegie Mellon University \\
\href{https://longvideohaystack.github.io}{\textbf{\texttt{https://longvideohaystack.github.io}}} \\[0.3em]
\scalebox{0.85}{    \href{https://huggingface.co/datasets/LVHaystack/LongVideoHaystack}{
    \includegraphics[height=1.0 em]{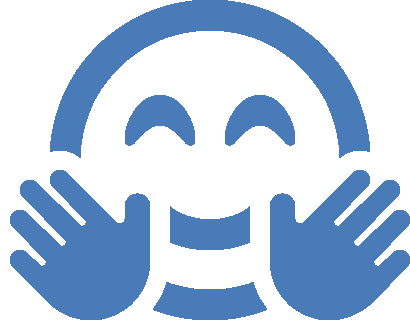}~\textbf{Data}} \quad
  \href{https://github.com/LongVideoHaystack/TStar}{\faGithub~\textbf{Code}} \quad
  \href{https://www.lvhaystackai.com/demo}{\faRocket~\textbf{Live Demo}} 
 \quad \href{https://lvhaystackai.com/assets/videos/Framework_0301_demo.mov}{\faVideo~\textbf{Video}} \quad \href{https://drive.google.com/drive/folders/1ig0XtZqGFYwERkARxCQMqIyKQjrtxcrx?usp=sharing}{\faFolderOpen~\textbf{Results}}
}
}

\begin{document}
\doparttoc 
\faketableofcontents 

\maketitle

\input{Sections/A0-Abstract}
\input{Sections/A1-Introduction}

\input{Sections/A2-Bench}
\input{Sections/A3-Methods}

\input{Sections/A4-Experiments}

\input{Sections/A5-Analysis}
\input{Sections/A7-RelatedWork}

\input{Sections/A8-conclusion}


{
    \small
    \bibliographystyle{ieeenat_fullname}
    \bibliography{main}
}

\clearpage
\appendix

\addcontentsline{toc}{section}{Appendix} 
\part{Appendix} 
\parttoc 

\input{Sections/A9-0-Appendix}


\end{document}

%% file: inc/inc_packages.tex
\usepackage{color}
\usepackage{epsfig}
\usepackage{graphicx}

\usepackage{booktabs}  
\usepackage{tabularx}               
\usepackage{ltablex}
\usepackage{tabularray}
\newcolumntype{C}{>{\centering\arraybackslash}X}
\usepackage{multirow}               
\usepackage{diagbox}                
\usepackage{hhline}                 
\usepackage{color}                  

\usepackage{extarrows}
\usepackage{makecell}
\usepackage{wrapfig}
\usepackage{colortbl}
\usepackage[dvipsnames]{xcolor} 
\usepackage{longtable}
\usepackage{textcomp}  
\usepackage{rotating} 
\usepackage{colortbl}    

\usepackage{adjustbox}
\usepackage{array}
\usepackage{floatflt}

\usepackage{amsmath}
\usepackage{bm}
\usepackage{nicefrac}
\usepackage{microtype}

\usepackage{changepage}
\usepackage{extramarks}
\usepackage{fancyhdr}
\usepackage{lastpage}
\usepackage{setspace}
\usepackage{soul}
\usepackage{xspace}
\usepackage[linesnumbered,ruled,vlined]{algorithm2e} 
\SetAlgoNlRelativeSize{-1} 
\SetAlgoVlined 
\SetCommentSty{textnormal}

\usepackage{enumitem}  

\usepackage{makecell}

\usepackage{pifont} 

\usepackage{algpseudocode}

\usepackage[symbol]{footmisc}

\usepackage[most]{tcolorbox}

\usepackage{caption}
\usepackage{scalefnt}

\usepackage{fontawesome5}

\usepackage{url}

\usepackage[toc,page,header]{appendix}
\usepackage{minitoc}

%% file: inc/inc_macros.tex
\newcolumntype{L}[1]{>{\raggedright\let\newline\\\arraybackslash\hspace{0pt}}m{#1}}
\newcolumntype{R}[1]{>{\raggedleft\let\newline\\\arraybackslash\hspace{0pt}}m{#1}}

\newcommand{\ignore}[1]{}

\makeatletter
\DeclareRobustCommand\onedot{\futurelet\@let@token\@onedot}
\def\@onedot{\ifx\@let@token.\else.\null\fi\xspace}

\makeatother

\newif\ifpropositionfirstitem
\propositionfirstitemtrue

\newdimen\abovecrulesep
\newdimen\belowcrulesep
\makeatletter
\patchcmd{\@@@cmidrule}{\aboverulesep}{\abovecrulesep}{}{}
\patchcmd{\@xcmidrule}{\belowrulesep}{\belowcrulesep}{}{}
\makeatother

\definecolor{mybluetitle}{HTML}{4B527E} 

\definecolor{codegreen}{HTML}{478058}
\definecolor{codegray}{rgb}{0.5,0.5,0.5}
\definecolor{codepurple}{HTML}{4F5E80} 
\definecolor{backcolour}{rgb}{0.95,0.95,0.92}
\lstdefinestyle{mystyle}{
    backgroundcolor=\color{backcolour},
    commentstyle=\color{codegreen},
    keywordstyle=\color{magenta},
    numberstyle=\tiny\color{codegray},
    stringstyle=\color{codepurple},
    basicstyle=\ttfamily\scriptsize,
    breakatwhitespace=false,
    breaklines=true,
    captionpos=b,
    keepspaces=true,
    frame=none,
    numbersep=5pt,
    showspaces=false,
    showstringspaces=false,
    showtabs=false,
    tabsize=2
}

\newtcolorbox{promptbox}[2][]{
    enhanced, 
    breakable,
    center title,
    left*=0pt, right*=0pt,
    boxsep=2pt, left=5pt, right=5pt,
    skin first=enhanced,
    skin middle=enhanced,
    skin last=enhanced,
    colback  = backcolour,
    fonttitle=\bfseries\rmfamily,
    fontupper=\scriptsize,
    title={\footnotesize\strut{#2}},
    #1
    }
\newcommand{\tcbcaption}[1]{
\noindent\begin{minipage}{0.5\textwidth}
\captionof{figure}{#1}
\end{minipage}
}

\newtcolorbox{onebox}[2][]{
    enhanced, 
    center title,
    left*=0pt, right*=0pt,
    boxsep=2pt, left=5pt, right=5pt,
    skin first=enhanced,
    skin middle=enhanced,
    skin last=enhanced,
    colframe = mybluetitle!90,
  colback  = mybluetitle!10,
    fonttitle=\bfseries\rmfamily\fontfamily{phv}\selectfont,
    title={\footnotesize\strut{#2}  \refstepcounter{subsubsection} \addcontentsline{toc}{subsubsection}{\string\numberline{\thesubsubsection}#2}
    },
    #1
    }

\definecolor{bggray}{HTML}{F5F5F5}
\definecolor{pvdblue}{HTML}{DAE8FC}
\definecolor{RoseQuartzBg}{HTML}{F7CAC9}
\definecolor{RoseQuartz}{HTML}{F5A798}
\definecolor{Serenity}{HTML}{92A8D1}
\definecolor{OrangeRed}{rgb}{1.0, 0.27, 0.0}
\definecolor{RoyalBlue}{cmyk}{1, 0.50, 0, 0}
\definecolor{Turquoise}{HTML}{0F4C81}
\definecolor{Mint}{rgb}{0.24, 0.71, 0.54}
\definecolor{Green}{rgb}{0.0, 0.120, 0.0}
\definecolor{MyBlue}{rgb}{0.46, 0.50, 0.61}
\definecolor{MyDarkBlue}{rgb}{0,0.08,0.8}
\definecolor{MyDarkGreen}{RGB}{45,155,45}
\definecolor{MyDarkRed}{rgb}{0.8,0.02,0.02}
\definecolor{MyOrange}{rgb}{1.0, 0.4, 0.2}
\definecolor{MyPurple}{RGB}{111,0,255}
\definecolor{MyRed}{rgb}{0.8,0.0,0.0}
\definecolor{MyGold}{rgb}{0.75,0.6,0.12}
\definecolor{MyDarkgray}{rgb}{0.66, 0.66, 0.66}
\definecolor{MyBrown}{rgb}{0.65, 0.16, 0.16}
\definecolor{MyMutedRose}{rgb}{0.58, 0.29, 0.35}
\definecolor{JiayuanColor}{rgb}{0.60,0.43,0.48}
\definecolor{erranColor}{rgb}{24, 40, 113}

\definecolor{citecolor}{HTML}{696FAD} 

\newcommand{\taskfull}{Long Video Haystack}
\newcommand{\task}{Temporal Search\xspace}
\newcommand{\taskshort}{TS\xspace}

\newcommand{\fancy}{\textit{T*}\xspace} 
\newcommand{\bench}{\textsc{LV-Haystack}\xspace}
\newcommand{\benchego}{\textsc{Haystack-Ego4D}\xspace}
\newcommand{\benchlv}{\textsc{Haystack-LVBench}\xspace}
\definecolor{darkgray}{gray}{0.9}
\newcommand{\llava}{LLaVA}

%% file: inc/math_commands.tex
\usepackage{amsmath,amsfonts,bm}

\def\eqref#1{equation~\ref{#1}}

\def\1{\bm{1}}

\DeclareMathAlphabet{\mathsfit}{\encodingdefault}{\sfdefault}{m}{sl}
\SetMathAlphabet{\mathsfit}{bold}{\encodingdefault}{\sfdefault}{bx}{n}



%% file: Sections/A0-Abstract.tex
\begin{abstract}

Efficiently understanding long-form videos remains a significant challenge in computer vision. In this work, we revisit temporal search paradigms for long-form video understanding and address a fundamental issue pertaining to all state-of-the-art (SOTA) long-context vision-language models (VLMs). Our contributions are twofold:
\textbf{First}, we frame temporal search as a \textbf{Long Video Haystack} problem -- finding a minimal set of relevant frames (e.g., one to five) from tens of thousands based on specific queries. Upon this formulation, we introduce \textbf{\bench}, the first dataset with 480 hours of videos, $15{,}092$ human-annotated instances for both training and evaluation aiming to improve temporal search quality and efficiency. Results on {\bench} highlight a significant research gap in temporal search capabilities, with current SOTA search methods only achieving $2.1\%$ temporal $F_1$ score on the \textsc{LongVideoBench} subset.

\textbf{Next}, inspired by visual search in images, we propose a lightweight temporal search framework, \textbf{\fancy} that reframes costly temporal search as spatial search. {\fancy} leverages powerful visual localization techniques commonly used in images and introduces an adaptive zooming-in mechanism that operates across both temporal and spatial dimensions. Extensive experiments show that integrating \fancy with existing methods significantly improves SOTA long-form video understanding. Under an inference budget of 32 frames, \textbf{T*} improves GPT-4o's performance from 50.5\% to \textbf{53.1\%} and LLaVA-OneVision-OV-72B's performance from 56.5\% to \textbf{62.4\%} on the \textsc{LongVideoBench} XL subset. Our code, benchmark, and models are provided in the Supplementary material.

\end{abstract}

%% file: Sections/A1-Introduction.tex
\begin{figure}[h!]
\centering
\includegraphics[width=0.45\textwidth]{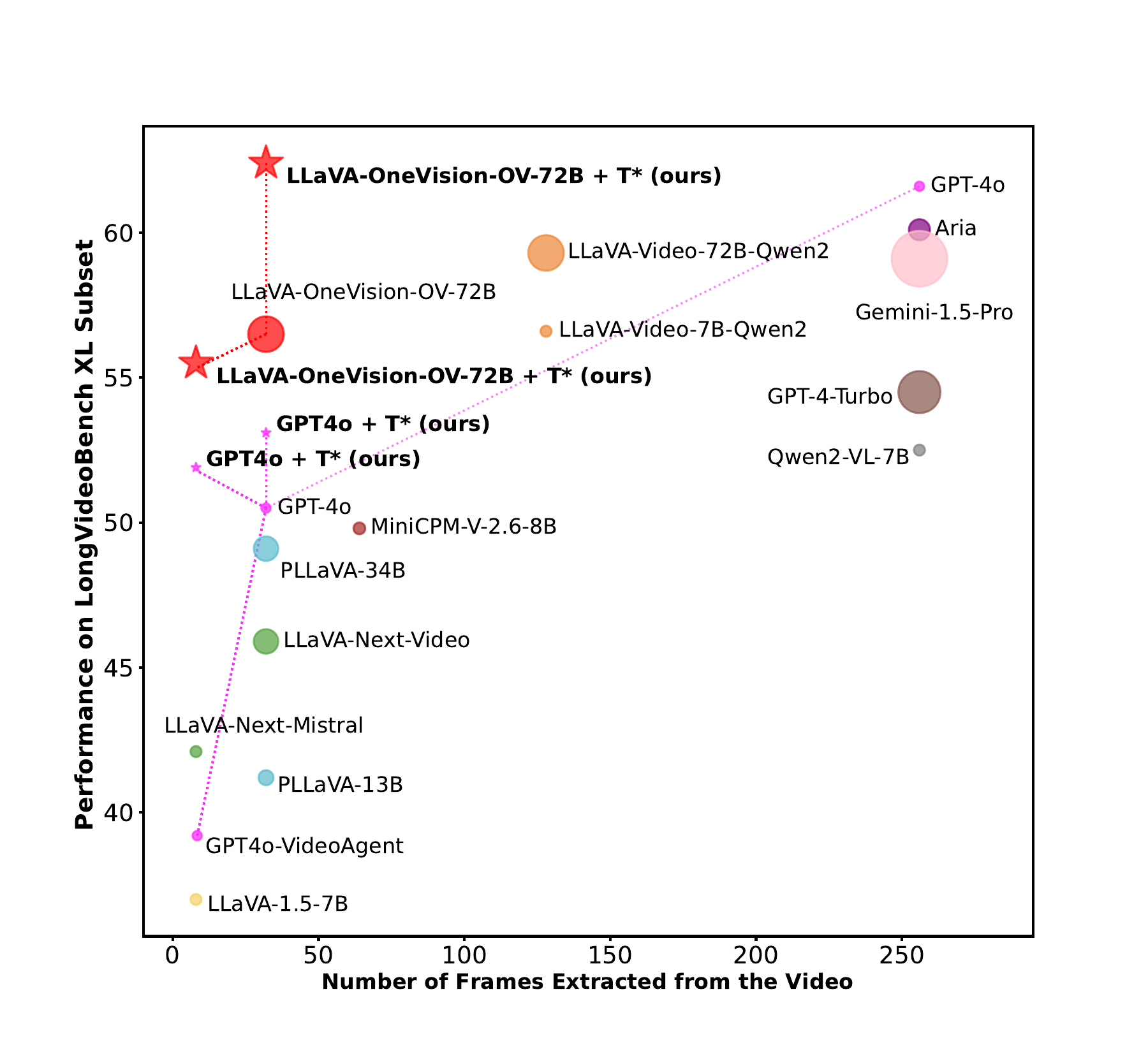}
\vspace{-0.2 cm}
\caption{
\textbf{
Long-form video understanding performance comparison
on LongVideoBench~\cite{wu2024longvideobench} XL subset (900-3600s)}. 
Open-sourced model size is indicated by marker size.
Our lightweight temporal search algorithm {\fancy} (\S\ref{sec:framework_overall}) improve SOTA models significantly: GPT-4o (50.5\%~$\rightarrow~$\textbf{53.1\%} and LLaVA-OneVision-OV-72B (56.5\%~$\rightarrow$~\textbf{62.4\%}), both with 32 frames.
}
\label{fig:gt_frames}
\end{figure}

\vspace{-0.1 em}
\section{Introduction}

As video understanding research expands from seconds-long to hour-long videos, ~\cite{wu2024longvideobench, hourvideo, ego4d}, video understanding tasks face fundamental challenges in quickly and accurately locating relevant frames in long-form videos~\cite{Liang2024, Di2024, Lei2018}.
Current large vision-language models (VLMs) often require a large number of tokens for frame processing, e.g., 576 tokens per image for LLaVA~\cite{liu2024visual} and Tarsier~\cite{wang2024tarsier}. This makes frame-by-frame analysis of long videos, which contain thousands of frames, computationally challenging to all state-of-the-art VLMs.
To overcome this challenge, temporal search \cite{wen2024too, wu2023discovering} has emerged as a fundamental paradigm, which is framed as a \textbf{Long Video Needle-in-a-Haystack} \cite{li2024needle, li2024end}: locating a minimal set of frames (needles) within thousands of frames from a long video (haystack) which is essential to answer the question.
Unlike traditional temporal localization \cite{Gao_2021_ICCV,lei2021detecting,yan2023unloc,ye2024improving,rodriguez2020proposal,zhang2020span,xiao2021boundary,tian2018audio,alwassel2021tsp,zhou2023adafocus} which identifies continuous temporal segments, temporal search focuses on selecting relevant frames across the entire video.

To this end, we introduce benchmark \textbf{\bench} specifically designed for temporal search on real-world long-form video. 
Unlike needle-in-a-haystack benchmarks \cite{2402.10790,2407.01437, wang2024needle,nelson2024needle, briakou2023searching,zhao2024needle,wang2024multimodal} using randomly inserted synthetic frames as ``needles'', {\bench} is built from real-world scenarios where humans answer questions by identifying few essential frames. We compile {\bench} using videos and questions from Ego4D (Egocentric videos, \citet{ego4d}) and LongVideoBench (Allocentric videos, \citet{wu2024longvideobench}), ensuring each question has an answer and a set of keyframes.
For LongVideoBench, we use keyframes and answers from the original dataset. For Ego4D, we annotate \textbf{15,092} QA instances from 988 videos spanning 423 hours with \textbf{45.7 million} frames, where each video lasts around 25 minutes with about 15 questions.
Furthermore, previous long-form video evaluations \cite{zhou2024mlvu,wang2024lvbench,wu2024longvideobench,li2024mvbench,rawal2024cinepile} primarily focus on task performance and overlook the evaluation of temporal search capabilities. We propose frame-centered temporal and visual metrics and derive frame-set similarity metrics like temporal and visual $F_1$ to compare model-selected and reference keyframes to evaluate search capabilities.

\begin{figure*}[h!]
\centering
\includegraphics[width=\textwidth]{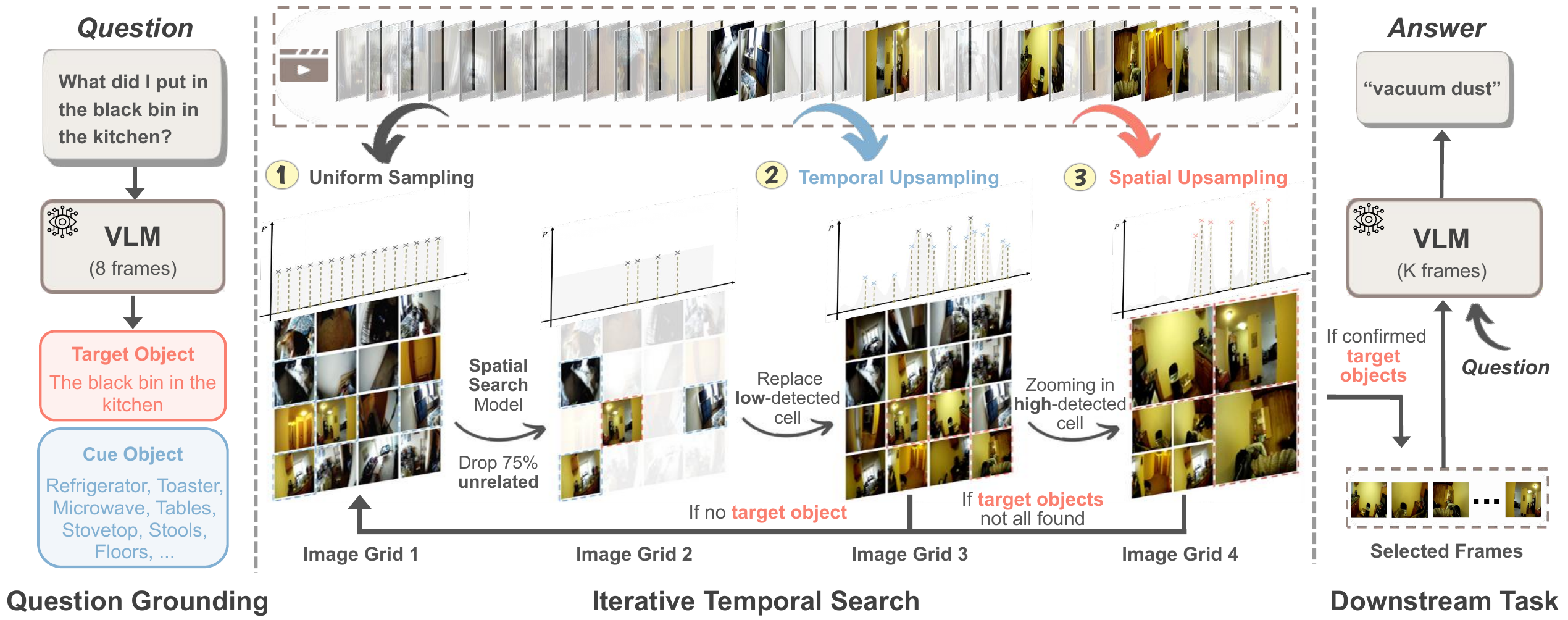}
\caption{\textbf{The {\fancy} framework that employs efficient temporal search for long-form video understanding.} {\fancy} employs an iterative temporal search approach to search keyframes essential to answer questions. Left: Question Grounding, where a visual language model identifies visual cues (target and cue object) from the textual question. Center: Iterative Temporal Search, formulated as Spatial Search where a spatial search model 
iteratively detects visual cues and upsamples relevant temporal/visual regions. Right: Downstream Task, where the visual language model answer questions using $K$ keyframes sampled from the final temporal search distribution as visual input.
}
\label{fig:framework}
\end{figure*}

Building upon the proposed benchmark, we examine the fundamental nature of temporal search in VLMs. Existing cluster-~\cite{wang2024videotree,jin2024chat,zhang2023simple,romero2024question,park2024too} or agent-based \cite{wang2024videoagent,yu2024self,fan2025videoagent,li2023videochat,wu2019adaframe} methods rely on costly frame-by-frame processing with VLM to identify keyframes.  
We draw inspiration from \textit{visual} search techniques like \textit{V*}~\cite{wu2024v}, which effectively conduct spatial search in Vision Transformers~\cite{minderer2022simple} in a coarse-to-fine manner, suggesting that \textit{temporal} search could be performed similarly. To unify temporal and spatial dimensions for video temporal search, we leverage the superior performance of image-language models over video-language models~\cite{hong2024cogvlm2visuallanguagemodels}, effectively recasting temporal search as a spatial search task.

Specifically, we propose \textbf{\fancy}, a temporal search framework reframed as a spatial search task by transforming frame sequences into a single large image, gradually refining temporal resolution by discarding irrelevant frames and inserting frames around key temporal regions. Acting like an agent, {\fancy} dynamically balances the spatial-temporal trade-offs by determining what spatial details to sacrifice, enhancing the temporal sampling probability in the promising time regions, and zooming-in images to achieve higher spatial resolution and recover details. This approach reduces search costs through multi-step zooming-in refinement, enabling more efficient and effective temporal search. With this unified approach, {\fancy} seamlessly integrates both temporal and spatial dimensions within a single image space to achieve efficient long-form video understanding.

Empirically, the iterative sampling-scoring-reweighting paradigm of {\fancy} results in $3\rm x$ computational efficiency in terms of FLOPs compared to frame-by-frame search. 
On long-form video understanding tasks, {\fancy} applied GPT-4o~\cite{2303.08774} and LLaVA-Onevision-OV-72B~\cite{li2024llava} achieves compatible performance while using $4\rm x$ fewer frames, outperforming pervious search and non-search methods. 
Furthermore, our fine-grained evaluation framework provides interpretable metrics for different components of video understanding, and our findings reveal that temporal search capabilities closely aligns with downstream performance.

%% file: Sections/A2-Bench.tex
\section{Temporal Search in Video Understanding}

To explore efficient temporal search with long-context VLMs, we formulate \task (\taskshort) similar to the needle-in-a-haystack \cite{wang2024multimodal} task, i.e., selecting few keyframes from the video to answer questions, which is critical for VLMs in processing long videos~\cite{VideoAgent, park2024too, liang2024end, wang2024videotree, tan2024koala, yu2024frame}.

\subsection{Task Formulation} 
\label{sec:formulation}

Given a video \( {V}=\{f_1, f_2, ..., f_N\} \) with $N$ frames and a question \( Q \), temporal search tries to find a minimal subset of \(k\) keyframes \( V^K = \{ f^{K}_1, f^{K}_2, \ldots, f^{K}_k \} \subseteq V \) that contains all critical information required to answer \( Q \). Specifically, the identified keyframe set requires two features:

\begin{itemize}
   \item \textbf{Completeness}:  $V^K$ should be a complete frame set to answer questions.  If the answer to $Q$ based on \({V}\) is $A$, then the answer derived from \(V^K\) should also be $A$.
   \item \textbf{Minimality}: \(V^K\) should contain only essential frames, with no redundant or irrelevant frames while maintaining completeness.
\end{itemize}

\subsection{The \textbf{\bench} Benchmark}
\label{sec:data_collection}

\input{Tables/dataset_statistics}

Based on this task formulation, we construct a benchmark specifically designed for Temporal Search. Each search instance in our dataset is represented as a tuple comprising four elements $\langle {V}, Q, V^K, A\rangle$, with a video ${V}=\{f_i\}_{i=0}^{N}$, a question $Q$, annotated keyframes $V^K=\{f^{K}_j\}_{j=0}^{k}$ and the answer $A$. 
Our benchmark consists of both egocentric and allocentric videos, sourced from Ego4D~\cite{ego4d} and LongVideoBench~\cite{wu2024longvideobench}, respectively.
For \benchego, we select video segments from the Ego4D NLQ validation set, with an average duration of 8.3 minutes per segment. These segments capture diverse scenarios such as object finding and shopping activities. We hire crowdworkers to identify the minimal set of keyframes required to answer task-specific questions and provide corresponding answers.
For \benchlv, we repurpose the LongVideoBench dataset for the temporal search task, where annotators verify and refine the original reference timestamps to ensure the frames contain minimal sufficient information to answer each question.
Statistics of our dataset can be found in Table~\ref{table:data-stats} and more data annotation details are listed in the Appendix~\ref{app:annotation_details}.

\subsection{Evaluation Metrics for Search Utility}
\label{sec:intrinsic_eval}
Our evaluation framework focuses on both search utility and efficiency. For search utility, we develop metrics comparing model-predicted keyframes with human annotations at both frame and set levels, addressing the challenge that multiple valid keyframe sets may exist for the same question.

\begin{algorithm*}[t]
\caption{Efficient Temporal Search with Dynamic Sampling}\label{alg_:Tstar}
\small
\SetKwInOut{Input}{Input}\SetKwInOut{Output}{Output}
\Input{Video $V$, target/cue objects $\{T,C\}$, keyframe count $K$, search budget $B$, threshold $\theta$}
\Output{Keyframes $F$ with timestamps $\tau$}

Initialize: $S,N \gets \mathbf{0}^L,\mathbf{1}^L$, $P \gets \frac{1}{L}\mathbf{1}^L$, $R \gets T$, $F,\tau \gets \emptyset$ \tcp*{$L=|V|$}

\While{$R \neq \emptyset$ \textbf{and} $B > 0$}{
    $I \gets \text{Sample}(P \odot N, g^2)$, $G \gets \text{Grid}(V[I])$, $B \gets B - g^2$ \tcp*{Sample and grid}
    $(C,O) \gets \text{Detect}(G)$ \tcp*{Get confidence maps and objects}
    
    \For{$i \in [1..|I|]$ \textbf{where} $O_i \cap R \neq \emptyset$}{
        $S[I_i],N[I_i] \gets C_i,0$ \tcp*{Update scores and mark visited}
        \If{$\text{Verify}(V[I_i]) > \theta$}{
            $F,\tau \gets F \cup \{V[I_i]\}, \tau \cup \{I_i/fps\}$, $R \gets R \setminus (O_i \cap R)$
        }
    }
    $P \gets \text{Normalize}(\text{Spline}(S,N))$ \tcp*{Update distribution}
}
\Return $\text{Sample}(V,S,K)$
\end{algorithm*}

\vspace{-7pt}
\paragraph{Frame-to-Frame Metrics.}
To evaluate alignment between a model-predicted frame \( f_{\text{pt}} \) and a human-annotated frame \( f_{\text{gt}} \), we consider two dimensions. \textbf{1) Temporal Similarity} measures the timestamp difference between \( f_{\text{pt}} \) and \( f_{\text{gt}} \), using a binary threshold to mitigate outlier effects. Two frames are considered similar if their temporal difference falls within this threshold. \textbf{2) Visual Similarity} adopts the Structural Similarity Index Measure (SSIM)~\cite{brunet2011mathematical} to identify the visual  similarity between the frames \( f_{\text{pt}} \) and \( f_{\text{gt}}\) based on structural details, luminance, and contrast.

\vspace{-7pt}
\paragraph{Set-to-Set Metrics.}
The major challenge in extending frame-to-frame metrics to frame set evaluation is defining what makes two sets \textit{similar}. We introduce \textbf{Precision} and \textbf{Recall} as two complementary metrics. Precision measures whether each model-selected frame aligns with at least one reference frame, while Recall evaluates whether reference frames are represented in the model's selection.

Let \( F_{\text{gt}} = \{f_{\text{gt}}^j\}_{j=1}^N \) denote the reference frame set and \( F_{\text{pt}} = \{f_{\text{pt}}^i\}_{i=1}^M \) represent the model-predicted frame set. We define precision and recall as follows:
\begin{equation}
\text{Precision}(F_{\text{pt}}, F_{\text{gt}}) = \frac{1}{|F_{\text{pt}}|}\sum_{f_{\text{pt}}^i \in F_{\text{pt}}} \text{sim}(f_{\text{pt}}^i, F_{\text{gt}}),    
\end{equation}
\begin{equation}
\text{Recall}(F_{\text{pt}}, F_{\text{gt}}) = \frac{1}{|F_{\text{gt}}|}\sum_{f_{\text{gt}}^j \in F_{\text{gt}}} \text{sim}(f_{\text{gt}}^j, F_{\text{pt}}),
\end{equation}
where \( \text{sim}(f^i, F') = \max\limits_{f^j \in F'} \text{sim}(f^i, f^j) \) defines the frame-to-set similarity for any frame and set. The \(\text{sim}\) function can measure either temporal or visual similarity. To balance search relevance (Precision) and coverage (Recall), we compute the $F_1$ score as the harmonic mean of them.

\subsection{Evaluation Metrics for Search Efficiency}
 
 \label{sec:searching-effi}

Previous research~\cite{VideoAgent,fan2025videoagent,wang2024videotree,wu2024v,park2024too} have primarily focused on downstream task performance and overlook the temporal search computational efficiency.
We evaluate search efficiency with three key metrics: \textbf{1) Frame Cost}, which measures the total number of frames processed, \textbf{2) FLOPs}, which quantifies the computational complexity, and \textbf{3) Latency}, which captures the total search time.

%% file: Tables/dataset_statistics.tex
\begin{table}[t]
\begin{center}
\scriptsize
\setlength\tabcolsep{1pt}
\setlength\extrarowheight{1pt}
\arrayrulecolor[gray]{0.7} 
\begin{adjustbox}{width=\linewidth}
\begin{tabular}{l|>{\raggedright\arraybackslash}l|l}
        \hline
        \rowcolor{gray!20}
        \textbf{Subset~ ~ ~ ~ ~} & \textbf{\textsc{Haystack-Ego4D}~ ~ ~ ~ ~ ~ ~} & \textbf{\textsc{Haystack-LVBench}~ ~ ~ ~ ~} \\
        \hline
        \textbf{Video Type} & Egocentric & Allocentric \\
        \rowcolor{gray!10}
        \textbf{\# video}         & 988 & 114 \\
        
        \textbf{\# length}        & 423 h    & 26.7 h  \\
        
                & - 25.7 min per video & - 14.1 min per video \\
        \rowcolor{gray!10}
        \textbf{\# frame}         & 45,700,000   & 2,200,000   \\
        \rowcolor{gray!10}
               & - 46,300 per video & - 19,100 per video \\
        \textbf{\# QA pair}       & 15,092 & 342 \\
                & - 15.3 per video & - 3.0 per video \\
        \rowcolor{gray!10}
        \textbf{\# keyframe}      & 28,300 & 496 \\
        \rowcolor{gray!10}
              & - 1.9 per question & - 1.5 per question \\
        \hline
\end{tabular}
\end{adjustbox}
\caption{Data Statistics of \bench.}
\vspace{-3 em}
\label{table:data-stats}
\end{center}
\end{table}

%% file: Sections/A3-Methods.tex
\section{\fancy: Efficient Temporal Search}
\label{sec:framework_overall}
{\fancy} facilitates long-form video understanding through temporal search, reformulated as spatial search with spatial search models. 
The framework (Figure~\ref{fig:framework}) comprises three phases: question grounding (\S\ref{subsec:grounding}), iterative temporal search (\S\ref{subsec:search}), and downstream task completion (\S\ref{subsec:downstream}). The first two phases conduct temporal search to identify keyframes, and the last phase forward these frames to a vision language model to answer questions.
The temporal search process is shown in Algorithm~\ref{alg_:Tstar}, and explained in detail as follows.

\input{Tables/2_3_search_perf}

\input{Tables/4_search_qa_perf}

\subsection{Question Grounding}
\label{subsec:grounding}
The question grounding phase aims to obtain target objects $T$ and cue objects $C$ essential for temporal search with spatial search models.
We sample $N$ frames at fixed intervals from video $V$, denoted as $\overline{V_N}$ for the VLM to scan. The VLM processes these frames with question $Q$ to identify two types of elements: (1) \textbf{Target Objects} $T$, visual elements directly relevant to answering the question,
(2) \textbf{Cue Objects} $C$, contextual elements indicating potential regions of interest.
These objects are formally represented as:
\vspace{-6pt}
\begin{equation}
\{T, C\} = \texttt{VLM}(\overline{V_N}, Q).
\vspace{-6pt}
\end{equation}
This query grounding phase identifies both primary targets and contextual cues helpful to answer the question, which are then used to guide the search process (\S\ref{subsec:search}). As shown in Figure~\ref{fig:framework}, for the question ``What did I put in the black dustbin?'', the VLM identifies both target object (dustbin) and cue objects (room corners, furniture placement).

\subsection{Iterative Temporal Search}
\label{subsec:search}

\paragraph{Initialization}
The temporal search begins by initializing a uniform probability distribution $P$ over frames, and a confidence threshold $\theta$ for object detection. The initial score distribution $S$ and non-visiting indicator $N_v$ are initialized as zero and one vectors over the total frame count $L$ (Algorithm~\ref{alg_:Tstar}, line 1). Each object $o \in \mathcal{O}$ is weighted at 1.0 for targets and 0.5 for cues to reflect search importance. The remaining target set $R$ starts with all targets $T$, while two empty sets $F$ and $\tau$ are created for storing keyframes and timestamps, respectively.

\vspace{-12pt}
\paragraph{Frame Sampling and Grid Construction}
In each iteration, the algorithm samples frames according to the current probability distribution $P$. We arrange sampled frames into a grid layout $G$ sized $g\times g$, where indices $I$ are first sampled and then used to construct the grid (Algorithm~\ref{alg_:Tstar}, line 3). The sampling process is defined as:
\vspace{-6pt}
\begin{equation}
    I = \text{Sample}(P \odot N_v, g^2)
\vspace{-6pt}
\end{equation}
where $g^2$ is the number of frames to sample and $\odot$ denotes element-wise multiplication. The search budget $B$ is then reduced by $g^2$ after grid construction.

\vspace{-12pt}
\paragraph{Object Detection and Scoring}
For each grid image, we perform object detection using a pre-trained model to identify both target and cue objects (line 4). The detection confidence for each grid cell $(i,j)$ is computed as:
\vspace{-6pt}
\begin{equation}
   C_{i,j} = \max_{o \in \mathcal{D}_{i,j}} (c_o \cdot w_o)
\vspace{-6pt}
\end{equation}
where $\mathcal{D}_{i,j}$ represents the detected objects in cell $(i,j)$, $c_o$ is the detection confidence, and $w_o$ is the object weight. When target objects are detected with sufficient confidence, they are added to the keyframe set and removed from the remaining targets (lines 5-8).

\vspace{-12pt}
\paragraph{Distribution Update}
The score distribution is updated by spline-based interpolation (line 9). For each sampled frame $f \in F_s$, we update its score and mark it as visited:
\vspace{-6pt}
\begin{equation}
   S_f = C_f, \quad N_{v,f} = 0
\vspace{-6pt}
\end{equation}
To capture temporal locality, we employ a window-based update for high-confidence frames:
\vspace{-6pt}
\begin{equation}
   S_{f\pm\delta} = \max(S_{f\pm\delta}, \frac{S_f}{|\delta| + 1}), \quad \delta \in [-w,w]
\vspace{-6pt}
\end{equation}
where $w$ is the window size. The probability distribution $P$ is then updated using spline interpolation and normalized.

The search process continues iteratively until either all target objects are found or the search budget $B$ is exhausted. Finally, the algorithm returns the top $K$ frames based on their final scores (line 10).

\subsection{Downstream Task Completion}
\label{subsec:downstream}
The final keyframes are selected using $\rm TopK$ operation on the score distribution (Algorithm~\ref{alg_:Tstar}, line 10), which returns $K$ frames with timestamps for downstream tasks, ensuring both relevance and temporal coverage.

%% file: Tables/2_3_search_perf.tex
\begin{table*}[th]
    \centering
    \setlength\tabcolsep{4pt} 
    \setlength\extrarowheight{1pt} 
    \arrayrulecolor[gray]{0.7} 
    \begin{adjustbox}{width=\linewidth}
    \begin{tabular}{l|c|c|c|c|c|c|c|c|c|c|c|c|c}
        \toprule
       \multirow{2}{*}{\bf Method} & \multirow{2}{*}{\textbf{Frames}$\downarrow$} & \multicolumn{6}{c|}{\bf \benchego} & \multicolumn{6}{c}{\bf \benchlv} \\  
       \cmidrule(lr){3-8} \cmidrule(lr){9-14}
       & & \multicolumn{3}{c|}{\textbf{Temporal}} & \multicolumn{3}{c|}{\textbf{Visual}} & \multicolumn{3}{c|}{\textbf{Temporal}} & \multicolumn{3}{c}{\textbf{Visual}} \\ 
       & & Precision $\uparrow$ & Recall $\uparrow$ & $F_1$ $\uparrow$& Precision $\uparrow$& Recall $\uparrow$& $F_1$$\uparrow$ & Precision $\uparrow$& Recall $\uparrow$& $F_1$ $\uparrow$& Precision $\uparrow$& Recall $\uparrow$& $F_1$$\uparrow$ \\
        \midrule
        \multicolumn{14}{l}{Baselines: Static Frame Sampling} \\ 
         \hline
         Uniform~\cite{wu2024longvideobench} & 8 & 1.0 & 3.4 & 1.6 & 58.0 & 63.0 & 60.2 & 1.4 & 6.3 & 2.2 & 56.0 & 72.0 & 62.7 \\
         Uniform~\cite{wu2024longvideobench} & 32 & 1.1 & 14.8 & 2.0 & 58.5 & 65.6 & 61.5 & 1.4 & 24.9 & 2.7 & 58.7 & 81.6 & 67.3 \\
         
        \midrule
        \multicolumn{14}{l}{Baselines: Adaptive Temporal Search} \\ 
        \hline
        VideoAgent~\cite{wang2024videoagent} & 10.1 & 1.7 & 5.8 & 2.7 & 58.0 & 62.4 & 59.9 & 1.2 & 8.5 & 2.1 & 58.8 & 73.2 & 64.7 \\
        Retrieval-based & 8 & 1.2 & 4.2 & 1.9 & 58.5 & 61.7 & 59.9 & 1.5 & 6.3 & 2.3 & 63.1 & 65.5 & 64.1 \\
        Retrieval-based & 32 & 1.0 & 13.8 & 1.9 & 58.5 & 65.4 & 61.4 & 1.3 & 21.8 & 2.4 & \textbf{59.9} & 80.8 & \textbf{67.8} \\

        \hline
        \rowcolor{gray!25}
        \multicolumn{14}{l}{Ours: $T^*$ for Zooming In Temporal Search} \\
        \rowcolor{gray!10}
        Attention-based & 8 & \underline{2.2} & \underline{7.5} & \underline{3.3} & 58.4 & \underline{62.5} & \underline{60.2} & 1.5 & 6.6 & 2.4 & \underline{63.6} & 68.6 & \underline{65.7} \\
        \rowcolor{gray!10}
        Training-based & 8 & 1.4 & 4.9 & 2.1 & 58.0 & 61.5 & 59.6 & 1.5 & 6.6 & 2.3 & 59.8 & 71.1 & 64.5 \\
        \rowcolor{gray!10}
        Detector-based & 8 & 1.7 & 5.8 & 2.7 & \underline{63.8} & 70.1 & 66.8 & \underline{1.6} & \underline{7.1} & \underline{2.5} & 58.4 & \underline{72.7} & 64.3 \\
        \rowcolor{gray!10}
        Detector-based & 32 & \textbf{1.8} & \textbf{26.3} & \textbf{3.4} & \textbf{{62.9}} & \textbf{76.2} & \textbf{68.9} & \textbf{1.7} & \textbf{28.2} & \textbf{3.1} & 58.3 & \textbf{83.2} & \textbf{67.8} \\
        
        \bottomrule
    \end{tabular}
    \end{adjustbox}
    \vspace{-0.2cm}
    \caption{
    Search utility results on {\bench}. 8-frame setting bests are \underline{underlined}, 32-frame setting bests are in \textbf{bold}. We show that more searched frames consistently improves recall but reduces precision in retrieval methods. Detector-based {\fancy} achieves best performance in 32-frame setting across metrics, demonstrating the effectiveness of visual grounding and iterative temporal search. Attention-based {\fancy} performs well in 8-frame setting but requires larger foundation models, thereby reducing efficiency.
        \vspace{-0.6 em}
    }
    \label{tab:main_bench_utility}
\end{table*}

\begin{table*}[th]
    \centering
    \setlength\tabcolsep{8pt} 
    \setlength\extrarowheight{1pt} 
    \arrayrulecolor[gray]{0.7} 
    \begin{adjustbox}{width=\linewidth}
    \begin{tabular}{l|c|c|c|c|c|c|c}
        \toprule
       \multirow{2}{*}{\bf Method} &  \multicolumn{4}{c|}{\bf Search Efficiency}  &   \multicolumn{3}{c}{\bf Overall Task Efficiency} \\  
          & Grounding & Matching & TFLOPs $\downarrow$ & Latency (sec) $\downarrow$ & TFLOPs $\downarrow$ & Latency (sec) $\downarrow$ & Acc $\uparrow$ \\
        \midrule
        \multicolumn{8}{l}{Baselines: Static Frame Sampling} 
         \\ 
         \hline
         Uniform-8~\cite{wu2024longvideobench} & N/A & N/A & N/A & 0.2 & 139.3 & 3.8 & 45.9  \\
        
        \midrule
        \multicolumn{8}{l}{Baselines: Adaptive Temporal Search} 
         \\ 
        \hline
        VideoAgent~\cite{wang2024videoagent} & GPT4$\times$4 & CLIP-1B$\times$840 & 536.5$^\dagger$ & 30.2 & 690.7$^\dagger$ & 34.9 & 49.2 \\
        Retrieval-based & N/A & YOLO-world-110M$\times$840 & 216.1 & 28.6 & 355.4 & 32.2 & 50.3 \\

        \rowcolor{gray!25}
        \multicolumn{8}{l}{Ours: $T^*$ for Efficient Temporal Search} \\

        \rowcolor{gray!10}
        Attention-based & \llava-72B$\times$3 & N/A & 88.9 & 13.7 & 228.2 & 17.3 & 49.6 \\
        \rowcolor{gray!10}
        Detector-based & \llava-7B$\times$1 & YOLO-world-110M$\times$49 & 33.3 & 7.5 & 172.6 & 11.1 & 50.8 \\
        \rowcolor{gray!10}
        Training-based & \llava-7B$\times$1 & YOLO-world-110M$\times$38 & \textbf{30.3} & \textbf{6.8} & \textbf{169.6} & \textbf{10.4} & \textbf{51.0}\\
        
        \bottomrule
    \end{tabular}
    \end{adjustbox}
    \vspace{-0.2cm}
    \caption{Efficiency results on the full \bench, including search efficiency and overall (search+downstream) efficiency. We report the search models used and their avg. call frequency (e.g., VideoAgent calls GPT-4 four times for grounding). 
    {\fancy} achieves high performance with significantly less computation and lower latency. VideoAgent's FLOPs ($^\dagger$) exclude GPT-4 costs due to its closed-source nature. All training and inference operations are carried out on a cluster of 8*H800 Nvidia GPUs.
    \vspace{-1.3 em}
    }
     
    \label{tab:main_bench_efficiency}
\end{table*}

%% file: Tables/4_search_qa_perf.tex
\begin{table*}[ht]
    \centering
    \setlength\tabcolsep{3pt} 
    \setlength\extrarowheight{1pt} 
    \arrayrulecolor[gray]{0.7} 
    \begin{adjustbox}{width=\linewidth}
    \begin{tabular}{l|c|c|c|c|c||l|c|c|c|c|c}
        \toprule
        \multicolumn{6}{c||}{\bf LongVideoBench} & \multicolumn{6}{c}{\bf Video-MME} \\
        \cmidrule{1-12}
        
        \multirow{3}{*}{\bf Model and Size} & \multirow{3}{*}{\bf \#Frame} & \multicolumn{4}{c||}{\textbf{Video Length}} & \multirow{3}{*}{\bf Model and Size} & \multirow{3}{*}{\bf \#Frame} & \multicolumn{4}{c}{\textbf{Video Length}} \\
        
        & & \textbf{XLong} & \textbf{Long} & \textbf{Medium} & \textbf{Short} & & & \textbf{Long} & \textbf{Medium} & \textbf{Short} & \textbf{Total} \\
        & & 15-60min & 2-10min & 15-60s & 8-15s & & & 41min & 9min & 1.3min & 17min \\

        \midrule
        
        GPT4o & 8 & 47.1 & 49.4 & 67.3 & 69.7 & GPT4o & 8 & 51.4 & 54.3 & 55.7 & 53.8 \\
        \rowcolor{gray!10}
         GPT4o + {\fancy} & 8 & \textbf{51.9} & \textbf{52.4} & \textbf{72.7} & \textbf{70.0} 
         & GPT4o + {\fancy} & 8 & \textbf{55.9} & \textbf{57.3} & \textbf{56.4} & \textbf{56.5} \\
         \arrayrulecolor{gray!40}\hline \arrayrulecolor{black}

        LLaVA-OneVision-72B & 8 & 53.7 & 57.4 & 74.1 & 73.0 & LLaVA-OneVision-72B & 8 & 52.6 & 55.5 & 59.6 & 55.9 \\
        \rowcolor{gray!10}
        LLaVA-OneVision-72B + {\fancy} & 8 & \textbf{55.5} & \textbf{63.7} & \textbf{76.3} & \textbf{73.5}
        & LLaVA-OneVision-72B + {\fancy} & 8 & \textbf{57.7} & \textbf{57.5} & \textbf{61.7} & \textbf{59.0} \\

        \midrule
        GPT4o & 32 & 50.5 & 57.3 & 73.5 & 71.4 & GPT4o & 32 & 56.3 & 60.7 & 68.3 & 61.8 \\

        \rowcolor{gray!10}
        GPT4o + {\fancy} & 32 & \textbf{53.1} & \textbf{59.4} & \textbf{74.3} & \textbf{71.4} & GPT4o + {\fancy} & 32 & \textbf{59.3} & \textbf{63.5} & \textbf{69.5} & \textbf{64.1} \\
        \arrayrulecolor{gray!40}\hline \arrayrulecolor{black}

        LLaVA-OneVision-72B & 32 & 56.5 & 61.6 & 77.4 & 74.3 & LLaVA-OneVision-72B & 32 & 60.0 & 62.2 & 76.7 & 66.3 \\

        \rowcolor{gray!10}
        LLaVA-OneVision-72B + {\fancy} & 32 & \textbf{62.4} & \textbf{64.1} & \textbf{79.3} & \textbf{74.6} & LLaVA-OneVision-72B + {\fancy} & 32 & \textbf{61.0} & \textbf{66.6} & \textbf{77.5} & \textbf{68.3} \\

        \midrule
        \color{gray} GPT-4o (0513) & \color{gray} 256 & \color{gray} 61.6 & \color{gray} {66.7} & \color{gray} 76.8 & \color{gray} 71.6 & \color{gray} Gemini-1.5-Pro (0615) & \color{gray} 1/0.5 fps$^{1*}$ & \color{gray} 67.4 & \color{gray} 74.3 & \color{gray} 81.7 & \color{gray} 75.0 \\
        \color{gray} Aria-8x3.5B & \color{gray} 256 & \color{gray} 60.1 & \color{gray} 64.6 & \color{gray} 76.6 & \color{gray} 69.4 & \color{gray} Qwen2-VL-72B & \color{gray} 768$^{3*}$ & \color{gray} 62.2 & \color{gray} 71.3 & \color{gray} 80.1 & \color{gray} 71.2 \\
        \color{gray} LLaVA-Video-72B-Qwen2 & \color{gray} 128 & \color{gray} 59.3 & \color{gray} 63.9 & \color{gray} 77.4 & \color{gray} 72.4 & \color{gray} GPT-4o (0615) & \color{gray} 384$^{2*}$ & \color{gray} 65.3 & \color{gray} 70.3 & \color{gray} 80.0 & \color{gray} 71.9 \\
        \color{gray} Gemini-1.5-Pro (0514) & \color{gray} 256 & \color{gray} 59.1 & \color{gray} 65.0 & \color{gray} 75.3 & \color{gray} 70.2 & \color{gray} LLaVA-Video-72B & \color{gray} 64 & \color{gray} 61.5 & \color{gray} 68.9 & \color{gray} 81.4 & \color{gray} 70.6 \\
        \color{gray} Qwen2-VL-7B & \color{gray} 256 & \color{gray} 52.5 & \color{gray} 56.7 & \color{gray} 67.6 & \color{gray} 60.1 & \color{gray} Aria-8x3.5B & \color{gray} 256 & \color{gray} 58.8 & \color{gray} 67.0 & \color{gray} 76.9 & \color{gray} 67.6 \\

        \bottomrule
    \end{tabular}
    \end{adjustbox}
    \caption{
Downstream task evaluation results show T* effectiveness as a temporal search module for VLMs on LongVideoBench and Video-MME (without subtitles for fair comparison). Using QA accuracy (\%) as the metric, we compare with top leaderboard models (shown in gray), noting these typically use substantially more frames, making direct comparisons challenging. Models are ranked by XLong video performance on LongVideoBench and total score on Video-MME, with frame counts indicated. All baseline figures are directly cited from their original publications. Standard deviations and more detailed analysis are available in Appendix~\ref{sec:app_c_analysis}. 
    }
    \label{tab:combined_table}    
\end{table*}

%% file: Sections/A4-Experiments.tex
\vspace{-5pt}
\section{Experimental Setup} 

\subsection{Evaluations on Search Utility and Efficiency} 
\label{sec:experimental_setup}

\paragraph{Datasets and models.}~~We evaluate on {\bench}  (Sec. \ref{sec:data_collection}). For downstream task efficiency evaluation, 8 searched frames are passed to LLaVA-OneVision-72B for all methods. We implement the spatial search model $H$ with three complementary ways: (1) attention-based using VLM's attention matrix,
(2) detector-based using object detector like YOLO-world \cite{cheng2024yolo},
(3) training-based using custom trained models. 
More details can be found in the codebase. 

\vspace{-14pt}
\paragraph{Evaluation Metrics.} 
We report search performance metrics from \S\ref{sec:intrinsic_eval} and \S\ref{sec:searching-effi} with a 5-second temporal threshold.

\vspace{-14pt}
\paragraph{Baselines.}
We include three representative sampling or search strategies:  \textbf{1) Uniform Sampling} following~\cite{wu2024longvideobench, li2024llava}; 
 \textbf{2) Temporal Search methods like VideoAgent}~\cite{wang2024videoagent} which leverages LLM-based video keyframe selection; \textbf{3) Retrieval-based methods} that score and rank all frames instead of {\fancy} search methods with iterative frame sampling.

\subsection{Evaluations on Downstream Tasks: Video QA} 

\paragraph{Datasets.}
We evaluate QA performance on a diverse set of video understanding tasks: LongVideoBench~\cite{wu2024longvideobench}, Video-MME~\cite{fu2024videomme}, EgoSchema~\cite{mangalam2023egoschema}, NExT-QA~\cite{nextqa} and Ego4D LongVideo QA, which we extended from Ego4D NLQ tast~\cite{grauman2022ego4d}. 

The videos range from brief clips (15 seconds) to extensive narratives (up to 60 minutes), covering tasks like temporal action reasoning, causal inference, and egocentric understanding. 
 
\vspace{-14pt}
\paragraph{Evaluation Metrics. }
We adopt accuracy on downstream QA tasks, following~\cite{zhou2024mlvu,wang2024lvbench,wu2024longvideobench,li2024mvbench,rawal2024cinepile}. 
\vspace{-14pt}
\paragraph{Baselines.} 
We test various open/closed-source VLMs, comparing {\fancy} and uniform frame selection with 8/32 frames. Implementation details can be found in Appendix~\ref{sec:Implementation}.

%% file: Sections/A5-Analysis.tex
\section{Experimental Results}
\subsection{Results on \textbf{\bench} Search Performance}

\input{Tables/5_nextqa_perf}

\input{Tables/6_ego_qa}

For search utility results (Table~\ref{tab:main_bench_utility}), attention-based {\fancy} achieves the best temporal metrics with higher precision, recall, and F1 scores with 8 frames. For 32 frames, detector-based {\fancy} has best performance across both temporal and visual metrics. Results show more frames improve recall but reduce precision in retrieval methods, demonstrating the effectiveness of visual grounding and iterative temporal search over retrieval-based methods or uniform sampling.

For search efficiency results, as shown in Table~\ref{tab:main_bench_efficiency}, {\fancy} achieve competitive accuracy with significantly fewer TFLOPs and lower latency than baselines. Training-based T* is particularly efficient (169.6 TFLOPs, 10.4s latency, 60.3\% accuracy) compared to VideoAgent (690.7 TFLOPs, 34.9s latency, 49.2\% accuracy). While uniform sampling has no search cost, it requires more frames to achieve similar performance, leading to more computational costs.

\subsection{Results on Downstream Tasks: Long Video QA}

We evaluate {\fancy} on four QA datasets by integrating it as a lightweight plugin into proprietary (GPT-4o) and open-source (LLaVA-OnVision-72B) vision-language models. For long-form videos, as shown in Table~\ref{tab:combined_table}, {\fancy} enhances VLM performance on LongVideoBench and VideoMME consistently across various frame budgets, video lengths, and VLMs. For short videos, Table~\ref{tab:main_nextqa} demonstrates that on NExT-QA and EgoSchema, {\fancy} outperforms other frame selection methods while using the least number of frames.

Notably, {\fancy} is particularly effective for longer videos and smaller frame budgets. On the LongVideoBench XLong subset with an 8-frame constraint, {\fancy} increases the SOTA models with a large margin, boosting GPT-4o performance from 47.1\% to 51.9\% with 8 frames, and LLaVA-
OneVision-OV-72B from 56.5\% to 62.4\% with 32 frames.

\subsection{Results on Ego4D LongVideo QA}

In the original Ego4D~\cite{grauman2022ego4d} NLQ task, each sample consists of a text query, an hours-long video, and a recommended video clip (approximately 10 minutes) that provides context for the query. In LVHaystack, we recruited seven annotators to answer the queries from the Ego4D NLQ test set and to generate corresponding answer options, resulting in a new long-video QA dataset from an ego-centric perspective. The dataset is partitioned into three subsets: tiny, dev, and test.

We report the performance of two mainstream open-source models, QWen2.5-VL~\cite{Qwen2.5-VL} and GPT4o. To facilitate future research, we provide results using both the full-length videos (hours-long video) and shorter clips (minutes-long video). The results are shown in Table~\ref{tab:main_ego4dqa}.

\section{Analysis}
\subsection{Time and Cost Complexity of {\fancy}}

{\fancy} can be viewed as a quaternary search guided by a heuristic informed by an object detector.
\textbf{In the worst case,} where the heuristic provides no useful information, {\fancy} randoms upsampling frames into the image grid with a complexity of $\mathcal{O}(\log L)$. 
\textbf{In the best case,} the heuristic always identifies the most relevant top cell in the grid. So {\fancy} operates as a quaternary search with a complexity of $\mathcal{O}(\log L)$.

Therefore, {\fancy} offers better efficiency compared to other methods that perform linear processing and examine all frames. 
And, the complexity for processing images in this scenario lies between \(\mathcal{O}(L)\) and
$   \mathcal{O}\left( \frac{\log L}{P} \right)$, 
where \( P \) is the probability that the object-detector-based heuristic selects the target frame from among image cells.

The computational overhead \( C \) of {\fancy} is:
\begin{equation}
C = \underbrace{N_{\text{VLM}} \cdot C_{\text{VLM}}}_{\text{Grounding Overhead}} + \underbrace{N_{\text{YOLO-world}} \cdot C_{\text{YOLO-world}}}_{\text{Matching Overhead}},
\end{equation}
Where, the overhead includes the frequency of reasoning and processing by the VLM and the cost associated with YOLO-world processing for each image grid. Empirical measurements of \( C_{\text{VLM}} \) and \( C_{\text{YOLO-world}} \) are provided in Section~\ref{sec:experimental_setup}.

\subsection{Sampling Iteration Dynamics}

\begin{figure*}[htbp]
    \centering
    \vspace{-0.6cm} 
    \includegraphics[width=1\linewidth]{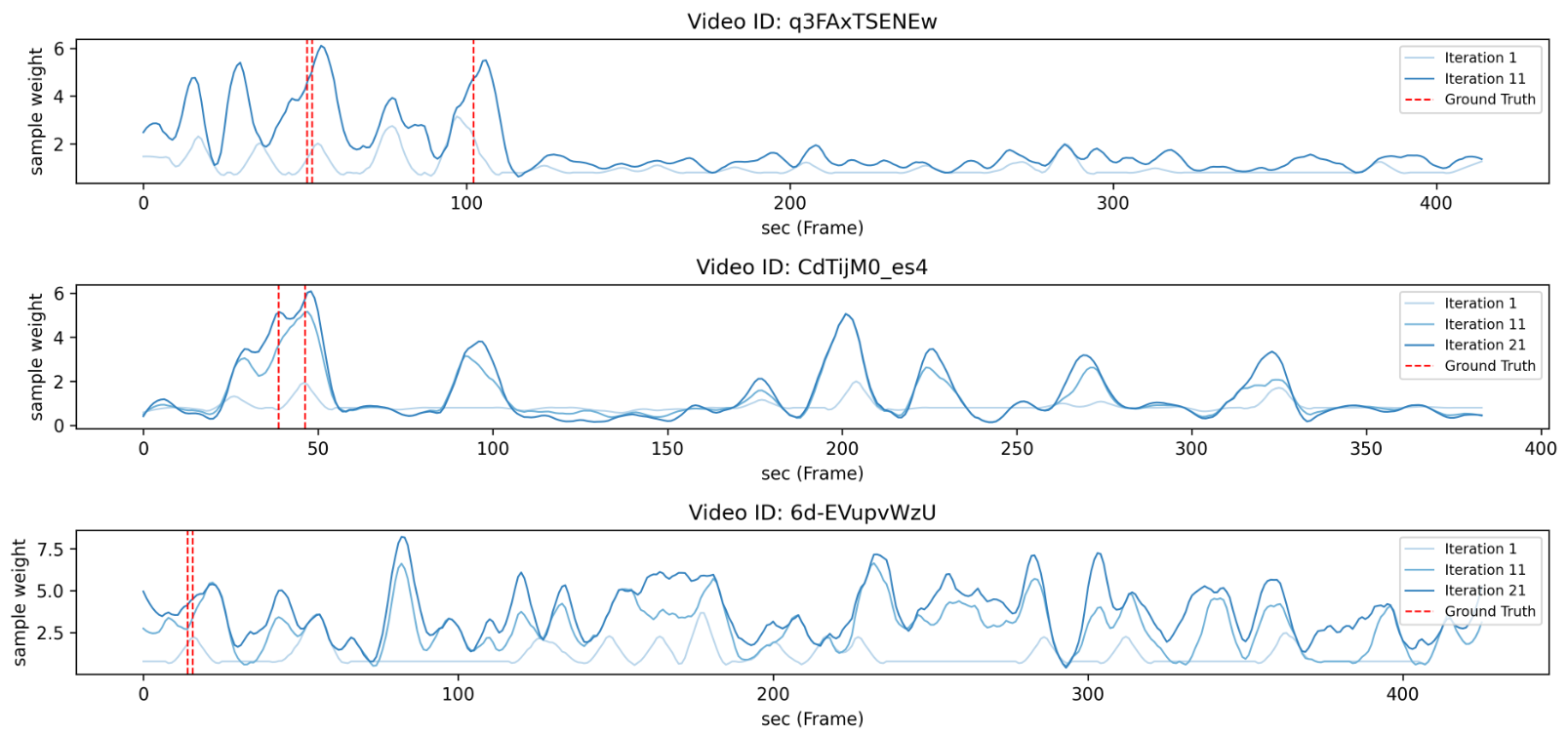} 
    \vspace{-10pt} 
    \caption{\textbf{Sampling weight dynamics over iterations for example videos.} Ground truth frames are marked in red. Sampling weights progressively focus on ground truth frames across iterations (1, 11, and 21), indicating improved model alignment with keyframes over time. Notably, due to the efficient sampling in temporal search, our model can simultaneously zoom in and focus on distantly located key frames (e.g., around 50s and 100s in the top plot).}
    \label{fig:revolution}
    \vspace{-5pt} 
\end{figure*}

We show the dynamics of temporal focus over multiple iterations of temporal search for three example videos in Figure~\ref{fig:revolution}. 
As one can observe, the results show that our method progressively aligns sampling weights with ground truth frames across iterations, enhancing the model’s ability to focus on relevant frames. Notably, even for distantly separated frames (e.g., around 50s and 100s in the top video), the model can simultaneously increase the sampling weights, demonstrating its ability to capture multiple critical frames in videos. This iterative refinement allows the model to identify and emphasize key frames accurately, improving overall long-form video understanding performance.

\subsection{Effect of Search Frame Count on Accuracy}

\begin{figure}[h!] 
\centering 
\includegraphics[width=\linewidth]{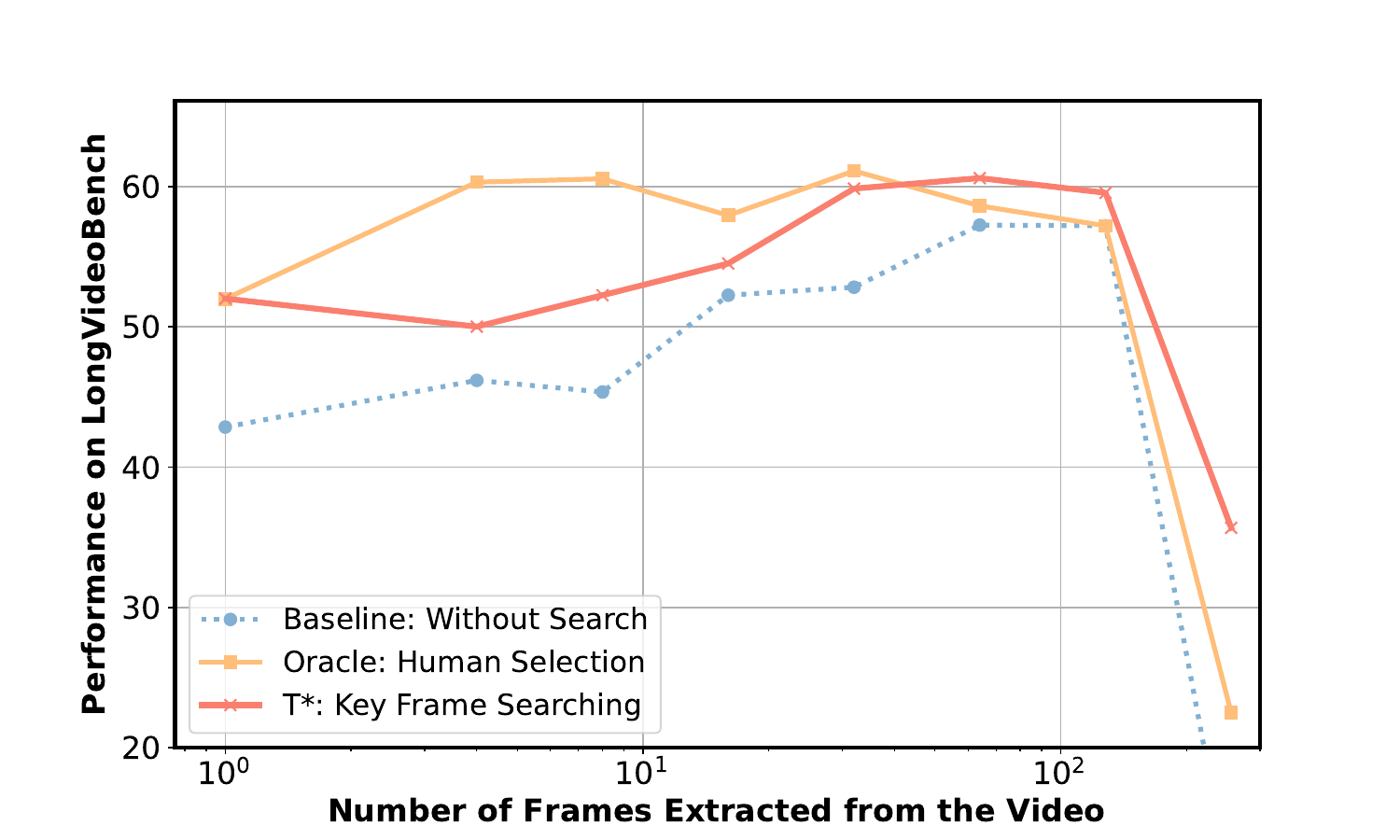} 
\caption{Performance improvement with increasing search frames.  {\fancy} consistently enhances accuracy and reaches near-human oracle performance at 64 frames.}
\label{fig:frames_vs_acc} 
\vspace{-0.5em}
\end{figure}

This section explores how the number of search frames influences the performance of our Visual Language Models (VLMs) on LongVideoBench.

Figure~\ref{fig:frames_vs_acc} illustrates the impact of different numbers of search frames on the performance of VLMs on LongVideoBench. The results show that {\fancy} consistently outperforms the baseline model across varying frame counts and closely approaches human selection (oracle) accuracy as the frame count increases. Notably, with 64 frames, {\fancy} achieves performance on par with human-selected frames, indicating that our method effectively captures the essential information with fewer frames.

%% file: Tables/5_nextqa_perf.tex
\begin{table}[ht]
    \centering
    \setlength{\tabcolsep}{3pt}
    \begin{adjustbox}{width=\linewidth}
    \setlength\tabcolsep{3pt}
    \setlength\extrarowheight{1pt}
    \begin{tabular}{l  | c  |c |c}
        \toprule
       \multirow{2}{*}{\bf Model} &  
      \multirow{2}{*}{\bf Frames }    &  \multicolumn{1}{c|}{\bf NExT-QA} &  \multicolumn{1}{c}{\bf EgoSchema} \\  
          & & \multicolumn{1}{c|}{ 0.7min} & \multicolumn{1}{c}{ 3min}  \\
          \hline

        \multicolumn{4}{l}{Baselines using Static Uniform Sampling} \\
        \hline
        InternVideo~\cite{wang2022internvideo} & 90 & 49.1 & 32.1\\
        MVU~\cite{ranasinghe2024understanding} & 16 & 55.2 & 60.3 \\

        LLoVi~\cite{zhang2023simple} & 90 & 67.7 & 57.6 \\
        LangRepo~\cite{kahatapitiya2024language} & 180 & 60.9 & 66.2\\
        \llava-OneVision-7B~\cite{li2024llava} & 32 & 79.4 & 65.4 \\
        \hline
          
        \multicolumn{4}{l}
        {Baselines using Adaptive Frame Selecting} \\
        \hline

        SeViLA~\cite{yu2024self} & 32 &  63.6 & 25.7 \\
        VideoAgent~\cite{VideoAgent} & 8.4 & 71.3 & 60.2\\

        LVNet~\cite{park2024too} & 12 &72.9 & 66.0\\
        VideoTree~\cite{wang2024videotree} & 63.2 & 73.5 & 66.2 \\

        VidF4~\cite{liang2024end} & 8 & 74.1 & -\\

        \hline
        \rowcolor{gray!25}
        \multicolumn{4}{l}{Ours: Plug in $T^*$ for Efficient Temporal Search} \\

        \llava-OneVision-7B~\cite{li2024llava} & 8 & 76.4 & 63.6 \\
        \rowcolor{gray!10}
        ~~+ {\fancy} & {8} &\textbf{80.4} & \textbf{66.6} \\

        \bottomrule
    \end{tabular}
    \end{adjustbox}
    \caption{Downstream task evaluation results by plugging in \fancy~as an additional temporal search method for VLMs on NExT-QA and EgoSchema. The video length is shorter than Table~\ref{tab:combined_table}. 
    Baseline results are directly cited from respective publications. 
    }
    \label{tab:main_nextqa}
    \vspace{-1 em}
\end{table}

%% file: Tables/6_ego_qa.tex
\begin{table}[ht]
    \centering
    \setlength{\tabcolsep}{3pt}
    \begin{adjustbox}{width=\linewidth}
    \setlength\tabcolsep{3pt}
    \setlength\extrarowheight{1pt}
    \begin{tabular}{l  | c  |cc | cc}
        \toprule
       \multirow{2}{*}{\bf Model} &  
      \multirow{2}{*}{\bf Frames }    &  \multicolumn{2}{c|}{\bf Tiny} &  \multicolumn{2}{c}{\bf Test} \\  
          & & \multicolumn{1}{c|}{ Clip} & \multicolumn{1}{c|}{ Video} & \multicolumn{1}{c|}{ Clip} & \multicolumn{1}{c}{ Video} \\
          \hline

        \multicolumn{6}{l}{Baselines using Static Uniform Sampling} \\
        \hline
        GPT4o & 8 & 45.5 & 41.5 & 45.9 & 45.3\\
        \rowcolor{gray!10}
        GPT4o + {\fancy} & {8} &{49.5} & {45.0} & 49.4 & 46.7\\

        \hline
        GPT4o & 32 & 49.0 & 45.5 & 52.3 & 51.0 \\
        \rowcolor{gray!10}
        GPT4o + {\fancy} & {32} &\textbf{{51.0}} & {{46.5}} & {54.9} & 52.5 \\
        
        \hline
        \hline
        QWen2.5VL 7B  & 8 & 37.0 & 32.0 & 40.0 & 38.8 \\

        \rowcolor{gray!10}
        QWen2.5VL 7B + {\fancy} & {8} & 38.5 & 37.0  & 42.7 & 40.3 \\
        \hline
        
        QWen2.5VL 7B  & 16 & 37.5 & 35.0 & 40.9 & 38.8  \\
        \rowcolor{gray!10}
        QWen2.5VL 7B + {\fancy} & {16} & 39.5 & 38.5  & 43.8 & 42.8 \\
        \hline
        \hline

        QWen2.5VL 72B & 8 & 45.0 & 45.0 & 51.0 & 50.1 \\
        \rowcolor{gray!10}
        QWen2.5VL 72B + {\fancy} & {8} &45.5 & 46.0  & 53.5 & 52.8\\
        \hline
        
        QWen2.5VL 72B & 16 & 49.0 & 49.5 & 53.6 & 50.6 \\
        \rowcolor{gray!10}
        QWen2.5VL 72B + {\fancy} & {16} & 50.0 & \textbf{50.0} & \textbf{55.1} &  \textbf{52.8}\\

        \bottomrule
    \end{tabular}
    \end{adjustbox}
    \caption{Downstream evaluation results on the \href{https://huggingface.co/datasets/LVHaystack/LongVideoHaystack}{Ego4D Longvideo QA dataset}. We extend the Ego4D NLQ task by including answer options and responses, and report performance for both clip-level and full video inputs using vision-language models.
    }
    
    \label{tab:main_ego4dqa}
    \vspace{-1 em}
\end{table}

%% file: Sections/A7-RelatedWork.tex
\section{Related Work}

\paragraph{Long-form Video Understanding. }
Recent attention mechanisms \cite{su2021roformer, beltagy2020longformer, zaheer2020big, dao2024flashattention, liu2024ringattention, tworkowski2023focused, li2024distflashattn, dao2022flashattention, liu2023ring,Zhou_2021_CVPR,10210078} and video transformers \cite{zhang2023video, lin2023video} improve temporal processing \cite{pang2021quality, peng2023yarn, liu2024deepseek,2409.10683} but struggle with long-range dependencies \cite{tay2020long}. Current solutions use compression \cite{song2024moviechat, jin2024chat} or frame selection \cite{ren2024timechat, gao2023mist, li2024llms, romero2024question, xu2023retrieval, yu2024self}, while existing benchmarks \cite{zhou2024mlvu, wang2024lvbench} focus on long videos, they mostly evlauate on downstream QA while we focus on temporal search evaluation.

\paragraph{Temporal Localization and Temporal Search.}
While temporal localization \cite{Gao_2021_ICCV,lei2021detecting,yan2023unloc,rodriguez2020proposal,zhang2020span,ye2023cross,xiao2021boundary,tian2018audio,alwassel2021tsp} struggles with boundary detection, recent keyframe selection advances from ``glance annotation" \cite{cui2022video} to caption-based \cite{kudo2023challenging} and fine-grained approaches \cite{lei2021detecting, chen2024verified}. Our work focus on a more challenging problem with longer videos.

\paragraph{Needle in a Haystack. }
Needle in a Haystack approaches span text \cite{2402.10790, 2407.01437} and multimodal \cite{briakou2023searching, zhao2024needle, wang2024multimodal, 2407.13766, dias2023keyframe} domains but rely on synthetic data, while our \taskfull~ focuses on real-world natural video contexts.

%% file: Sections/A8-conclusion.tex
\section{Conclusion}

In this work, we revisit temporal search paradigms for long-form video understanding, studying a fundamental issue pertaining to all SOTA long-context vision-language models (VLMs).
First, we formulate temporal search as a Long Video Haystack problem, i.e., finding a minimal set of relevant frames among tens of thousands of frames from real-world long videos given specific queries. 
To validate our formulation, 
we create \textbf{\bench}, the first benchmark containing $15{,}092$ 
human-annotated
instances with a set of fine-grained evaluation metrics for assessing temporal search quality and computational efficiency. 
Empirical results on \textbf{\bench} under SOTA temporal search methods reveal a significant gap in temporal search capabilties.
Next, we re-think temporal search in long-form videos and propose a lightweight temporal search framework, \textbf{\fancy}, which casts the expensive temporal search as a spatial search problem.
Extensive experiments show that when integrated with video understanding models, \textbf{\fancy} significantly improves SOTA performance.
We hope that \textbf{\bench} and {\fancy} framework will drive meaningful advancements in developing efficient long-form video understanding systems.

\section*{Limitations}
A potential limitation of our work lies in the assumption that most problems can be addressed with a few keyframes, which may not fully extend to more complex tasks requiring a broader context or dense reasoning. Additionally, our approach focuses primarily on visual cues, without leveraging other modalities such as audio or subtitles, which can be explored in future work to enhance multi-modal understanding.


%% file: Sections/A9-0-Appendix.tex
\newpage
\input{Sections/A9-1-Ablations}

\input{Sections/A9-2-Complexity-Analysis}

\input{Sections/A9-3-Detail-Analysis}

\input{Sections/A9-4-Implementation-Details}

\begin{figure*}[!t]
    \centering
    \includegraphics[width=0.8\linewidth]{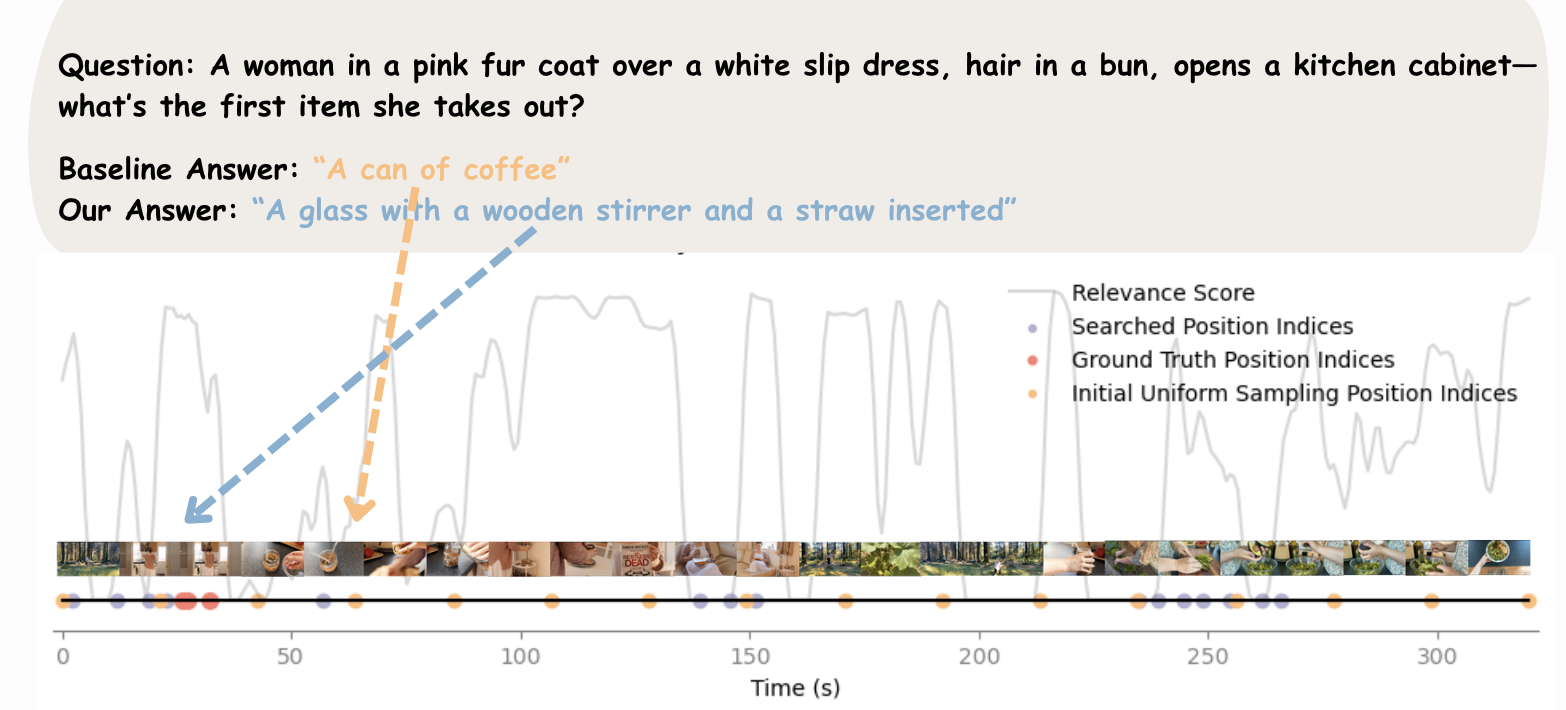}
    \caption{
       The visualization of frame selection results demonstrates the effectiveness of our approach compared to baseline methods. Our method consistently identifies more relevant and temporally diverse keyframes, capturing important frames that directly address the question. In contrast to baseline approaches which may select redundant or less informative frames, our strategy achieves better coverage of key events while maintaining temporal coherence across the video sequence. 
    }
    \label{fig:case_study}
\end{figure*}

\input{Sections/A9-5-Annotation}

\begin{figure*}[t] \centering \includegraphics[width=0.68\linewidth]{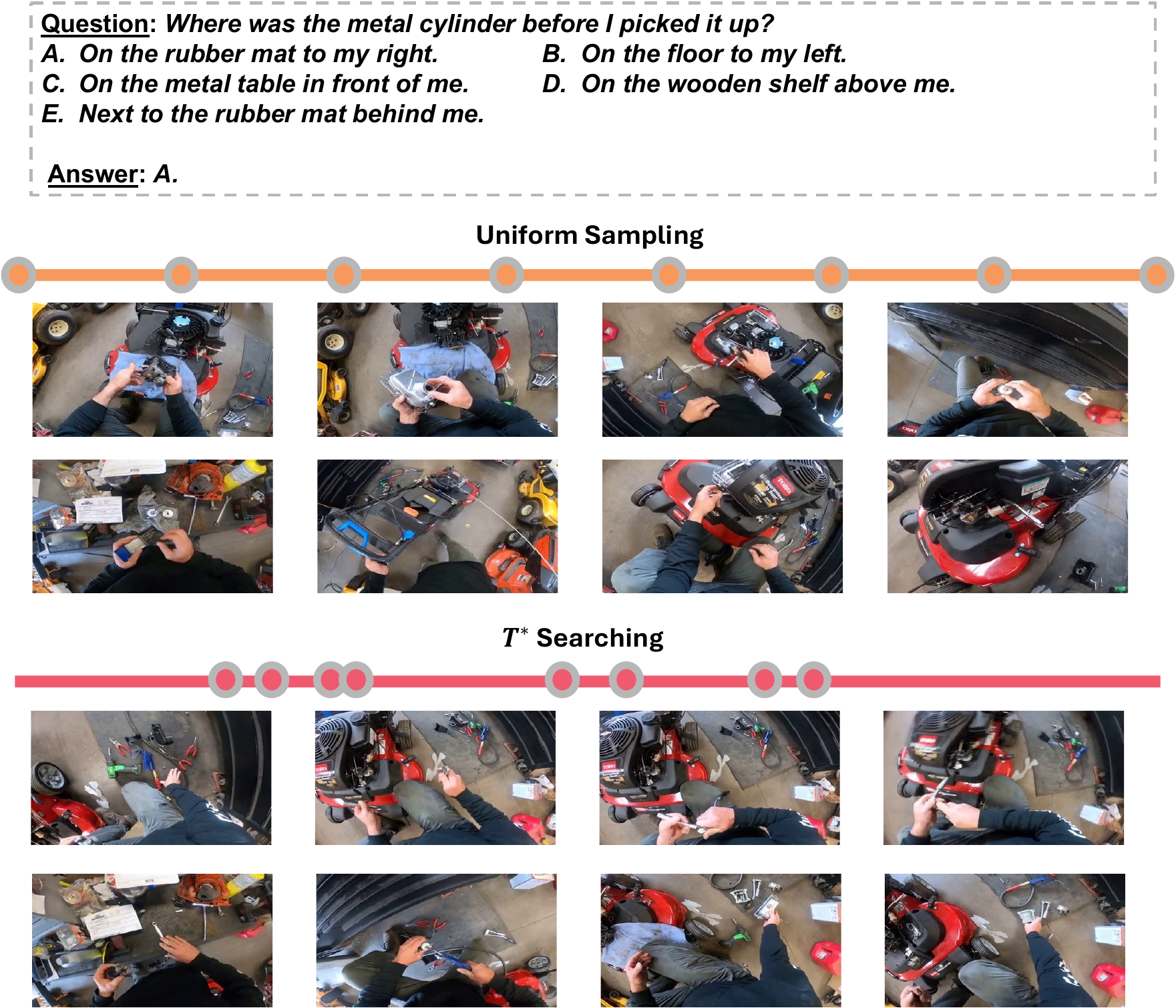} \caption{ \textbf{Comparison of uniform sampling and T* sampling for long-format video understanding.} In this example, the task involves identifying a ``metal cylinder'' in an hour-long video. Uniform sampling fails to include relevant frames, as it randomly selects 8 frames across the video. In contrast, T* sampling dynamically selects frames containing the metal cylinder, providing the necessary visual context for effective understanding. } \label{fig:sampling_case} \end{figure*}

\input{Sections/A9-6-Case_Statement}

\input{Sections/A9-7-Prompt-Statement}

\input{Sections/A9-8-Code}
\input{Sections/A9-9-Limitations-and-Broader-Impact}

%% file: Sections/A9-1-Ablations.tex
\vspace{40pt}

\section{Ablation Study}

This section investigates the sensitivity of different parameters in our proposed {\fancy} framework.

\subsection{Ablation on Question Grounding}

\textbf{Question Grounding} transforms the original question into spatially queryable targets, as detailed in Section 4.
In this study, we examine how increasing computational resources for Question grounding affects the efficiency and quality of the search process.
Specifically, we analyze the impact of scaling Vision-Language Models (VLMs) from 7B (LLaVA) to 72B (LLaVA) parameters, as well as varying the number of initial frames, from 8 to 32.


The results, summarized in Table~\ref{tab:abl_reasoning}, indicate that increasing the VLM size and the number of initial frames marginally enhances both the effectiveness of the search process and downstream task performance. 
These results suggest that Question Grounding can be effectively achieved with modest resource allocations, offering a favorable balance between performance and resource usage.

\input{Tables/result_ablation_reasoning}

\subsection{Ablation on Searching Algorithm}

{\fancy} aims to reduce computational overhead by partially representing the video as an \(n \times n\) image grid. This approach leverages well-trained image models to systematically replace irrelevant grid cells until the target is found, based on a specified threshold \(\theta\).

\begin{figure}[h!]
\centering
\includegraphics[width=0.45\textwidth]{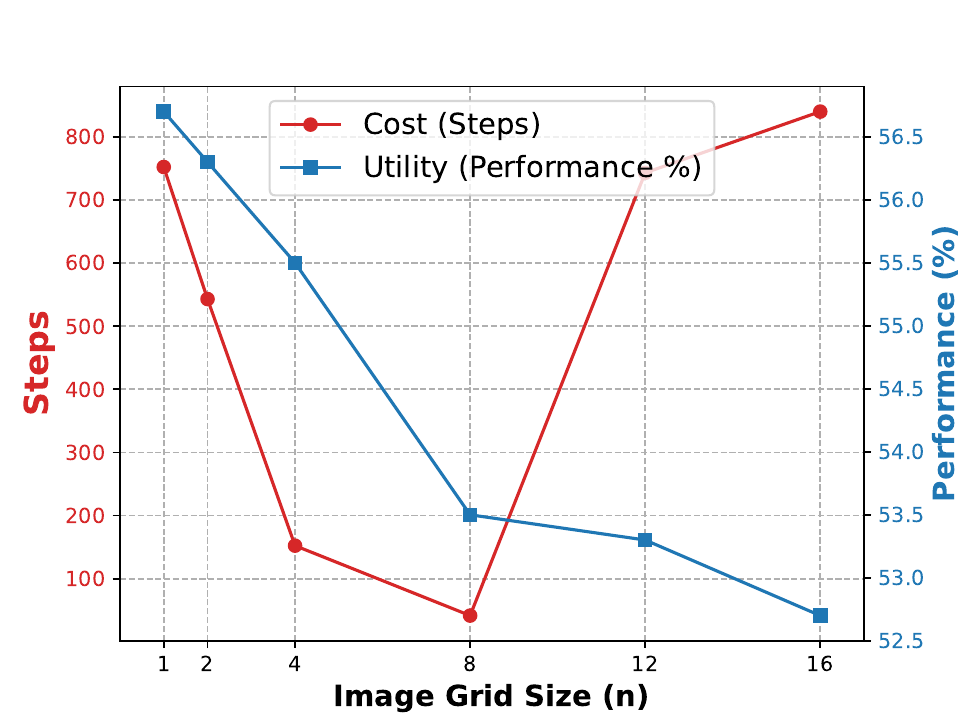}
\caption{\textbf{Grid Size Impact on Search Performance}. The red line represents the average number of search iterations for different image grid configurations, while the blue line shows the performance on the LongVideoBench~\cite{wu2024longvideobench} XL subset using 8 frames and the \llava-72B as the downstream QA model.}
\label{fig:grid_size_impact}
\vspace{-1em}
\end{figure}

\noindent
\textbf{Impact of Grid Size \((n)\)}:
 We investigate how the configuration of the concatenated image grid affects both the search cost and the efficacy of the search process. Figure~\ref{fig:grid_size_impact} displays the impact of varying grid sizes \(n\) (represented on the X-axis) on the average number of search steps and the corresponding average accuracy on LongVideoBench~\cite{wu2024longvideobench} XL subset. 


\begin{figure}[h!]
\centering
\includegraphics[width=0.45\textwidth]{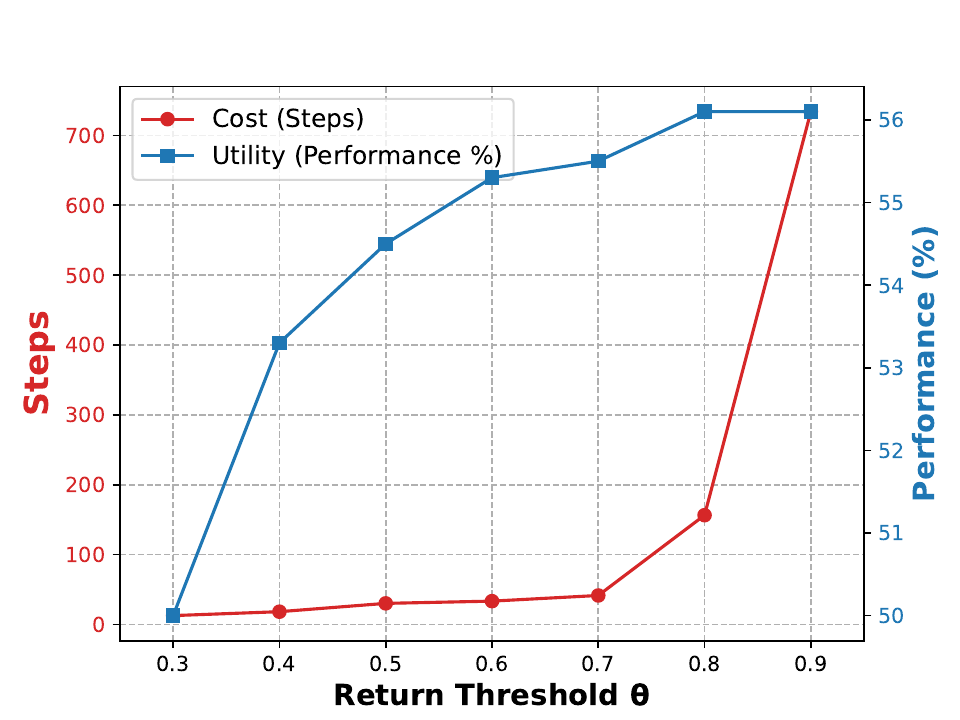}
\caption{\textbf{Impact of Return Threshold \(\theta\)}. The graph clearly illustrates the trade-off between threshold settings and search performance: lower thresholds result in quicker searches but may reduce accuracy, while higher thresholds enhance accuracy at the expense of increased search steps and computational cost.}
\label{fig:threshold_impact}
\vspace{-1em}
\end{figure}

\noindent
\textbf{Impact of Return Threshold \(\theta\)}: We examine how varying the return threshold \(\theta\) impacts the efficiency and efficacy of the search process. As demonstrated in Figure~\ref{fig:threshold_impact}, increasing the threshold tends to improve the accuracy of the search results but at the cost of increased computational effort. This trade-off is critical; thus, we have selected a default threshold of \(\theta = 0.6\) for a balanced approach.

\subsection{Ablation on Searching Utility Metrics} \label{app:semantic}

In our primary evaluation framework, we adopt Temporal Similarity and Visual Similarity as core metrics for measuring search utility. To further investigate the robustness of our framework, we include semantic distance as an additional metric for ablation studies. Semantic distance measures the alignment of high-level features between predicted and annotated frames, as encoded in pretrained models such as~\texttt{openai/clip-vit-large-patch14}. 

The results on \textsc{Haystack-Ego4D} are shown Appendix Table~\ref{tab:semantic-distance}. While this metric provides insights into the semantic relevance of frames, our results reveal that its scores are closely clustered across methods, ranging from 87.9 to 89.2. Therefore, semantic distance, while informative, does not significantly discriminate between methods due to the high-level feature similarities shared across retrieved keyframes. We exclude it as an evaluation metric to maintain focus on the more distinctive temporal and visual search utility.

\begin{table*}[th]
    \centering
    \setlength\tabcolsep{9pt} 
    \setlength\extrarowheight{3pt} 
    \arrayrulecolor[gray]{0.7} 
    \begin{adjustbox}{width=\linewidth}
    \begin{tabular}{l|c|c|c|c|c|c|c|c|c|c}
        \toprule
       \multirow{2}{*}{\bf Method} & \multirow{2}{*}{\textbf{Frames}$\downarrow$} & \multicolumn{9}{c|}{\bf \benchego}  \\  
       \cmidrule(lr){3-8} \cmidrule(lr){9-11} 
       & & \multicolumn{3}{c|}{\textbf{Temporal}} & \multicolumn{3}{c|}{\textbf{Visual}} & \multicolumn{3}{c}{\bf Semantic}\\ 
       & & Precision $\uparrow$ & Recall $\uparrow$ & $F_1$ $\uparrow$ & Precision $\uparrow$ & Recall $\uparrow$ & $F_1$ $\uparrow$ & Precision $\uparrow$ & Recall $\uparrow$ & $F_1$ $\uparrow$  \\
        \midrule
        \multicolumn{11}{l}{Baselines: Static Frame Sampling} \\ 
         \hline
         Uniform~\cite{wu2024longvideobench} & 8 & 1.0 & 3.4 & 1.6 & 58.0 & 63.0 & 60.2 & 87.2 & 89.3 & \underline{88.2} \\
         Uniform~\cite{wu2024longvideobench} & 32 & 1.1 & 14.8 & 2.0 & 58.5 & 65.6 & 61.5 & \textbf{87.3} & 90.4 & 88.8 \\
         
        \midrule
        \multicolumn{11}{l}{Baselines: Adaptive Frame Selection} \\ 
        \hline
        VideoAgent~\cite{wang2024videoagent} & 10.1 & 1.7 & 5.8 & 2.7 & 58.0 & 62.4 & 59.9 & 87.0 & 88.9 & 87.9 \\
        Retrieval-based & 8 & 1.2 & 4.2 & 1.9 & 58.5 & 61.7 & 59.9 & 87.3 & 88.7 & 88.0 \\
        Retrieval-based & 32 & 1.0 & 13.8 & 1.9 & 58.5 & 65.4 & 61.4 & \underline{87.3} & 90.5 & 88.9 \\

        \hline
        \rowcolor{gray!25}
        \multicolumn{11}{l}{Ours: $T^*$ for Zooming In Temporal Search} \\
        \rowcolor{gray!10}
        Attention-based & 8 & \underline{2.2} & \underline{7.5} & \underline{3.3} & 58.4 & \underline{62.5} & \underline{60.2} & 87.3 & \underline{89.1} & 88.1 \\
        \rowcolor{gray!10}
        Training-based & 8 & 1.4 & 4.9 & 2.1 & 58.0 & 61.5 & 59.6 & 87.2 & 89.0 & 88.0 \\
        \rowcolor{gray!10}
        Detector-based & 8 & 1.7 & 5.8 & 2.7 & \underline{63.8} & 70.1 & 66.8 & 87.2 & 88.9 & 87.9 \\
        \rowcolor{gray!10}
        Detector-based & 32 & \textbf{1.8} & \textbf{26.3} & \textbf{3.4} & \textbf{62.9} & \textbf{76.2} & \textbf{68.9} & 87.2 & \textbf{91.4} & \textbf{89.2} \\
        
        \bottomrule
    \end{tabular}
    \end{adjustbox}
    \vspace{-0.2cm}
    \caption{
Results of searching utility on {\bench}. Best results for the 8-frame setting are \underline{underlined}, and best results for the 32-frame setting are in \textbf{bold}. We include semantic metric (detailed in Appendix~\ref{app:semantic}) for ablation. Scores range closely from 87.9 to 89.2, showing limited differentiation across methods compared to temporal and visual metrics.
    }
    \label{tab:semantic-distance}
    \vspace{-1 em}
\end{table*}

\subsection{Correlation between Search Utility and Video Understanding}

As discussed in Section~\ref{sec:intrinsic_eval}, we propose multiple temporal and visual metrics to evaluate search utility. To identify the metrics most correlated with long-form video understanding, we analyzed the Pearson and Spearman correlation coefficients between utility scores and downstream task accuracy.
Table~\ref{tab:correlation_summary} shows that Temporal $F1$ has the highest Pearson correlation, while Temporal Precision has the highest Spearman correlation with downstream performance, highlighting these metrics as strong predictors of effective video understanding.

\input{Tables/result_correlation}

%% file: Tables/result_ablation_reasoning.tex
\begin{table}[th!]
    \centering
    \setlength{\tabcolsep}{3pt}
    \begin{adjustbox}{width=\linewidth}
    \setlength\tabcolsep{3pt}
    \setlength\extrarowheight{1pt}
    \begin{tabular}{l  | c  |c |c | c }
        \toprule
       \multirow{1}{*}{\bf Grounding VLM} &  
      \multirow{1}{*}{\bf Frames }    &  \multicolumn{1}{c|}{\bf TFLOPs} &  \multicolumn{1}{c|}{\bf  Visual F1 } &  \multicolumn{1}{c}{\bf  QA Acc }\\  
          \toprule
        \llava-OneVision-7B & {8} &{26.9} & {59.9} & 59.8\\
        \midrule
        \llava-OneVision-72B & {8} &{148.5} & {60.6} & 59.9\\
        \midrule

        \llava-OneVision-7B & {32} &{108.2} & {60.7} & 60.3\\
        
        \bottomrule
    \end{tabular}
    \end{adjustbox}
    \caption{Impact of VLM size and initial frame count on question grounding and search effectiveness on {\bench}. Experimental settings are aligned with the baseline setup reported in Section 4 (main paper).
    Our results indicate that increasing the resources for question grounding results in marginal improvements in search effectiveness ($<1$\% gain in QA Acc.).
    }
    \label{tab:abl_reasoning}

    \vspace{-1 em}
\end{table}

%% file: Tables/result_correlation.tex

\begin{table}[ht]
    \centering
    \begin{minipage}{\linewidth} 
        \centering
        \small 
        \setlength{\tabcolsep}{2pt} 
        \renewcommand{\arraystretch}{0.9} 
        \begin{tabular}{lcccc}
            \toprule
            \textbf{Metric} & \scriptsize \makecell{\textbf{Pearson} \\ \textbf{Correlation}} & \scriptsize \makecell{\textbf{Pearson} \\ \textbf{p-value}} & \scriptsize \makecell{\textbf{Spearman} \\ \textbf{Correlation}} & \scriptsize \makecell{\textbf{Spearman} \\ \textbf{p-value}} \\
            \midrule
            Temporal $F_1$       & \textbf{0.901} & 0.037 & 0.700 & 0.188 \\
            Temporal Precision & 0.828 & 0.084 & \textbf{0.975} & 0.005 \\
            Visual $F_1$         & 0.829 & 0.083 & 0.600 & 0.285 \\
            Temporal Recall   & 0.655 & 0.231 & 0.700 & 0.188 \\
            Visual Recall     & 0.568 & 0.317 & 0.500 & 0.391 \\
            Visual Precision     & 0.327 & 0.591 & 0.100 & 0.873 \\
            \bottomrule
        \end{tabular}
        \caption{ Pearson and Spearman correlations (with p-values) between search utility metrics and downstream task accuracy. The highest correlations are highlighted in \textbf{bold} for Temporal $F_1$ and Temporal Precision, suggesting they are strong predictors of effective video understanding performance.    }
        \label{tab:correlation_summary}
    \end{minipage}
    \vspace{-1 em}
\end{table}

%% file: Sections/A9-2-Complexity-Analysis.tex
\section{Complexity Analysis on {\fancy}}
\label{sec:complexity}

In this section, we analyze the time and cost complexity of the {\fancy} search algorithm. {\fancy} leverages adaptive temporal and spatial upsampling to efficiently collect partial information from the video and progressively determine the keyframe distribution. Based on previous observations, {\fancy} prioritizes high-probability regions for efficient keyframe localization, similar to an A* search algorithm. By retaining only a portion of the video grid cells in each iteration (up to \(1/b\) of total cells), {\fancy} effectively performs a multi-branch search guided by a heuristic scoring function, forming a \(b\)-way search tree.

\begin{figure}[h!]
\centering
\includegraphics[width=0.45\textwidth]{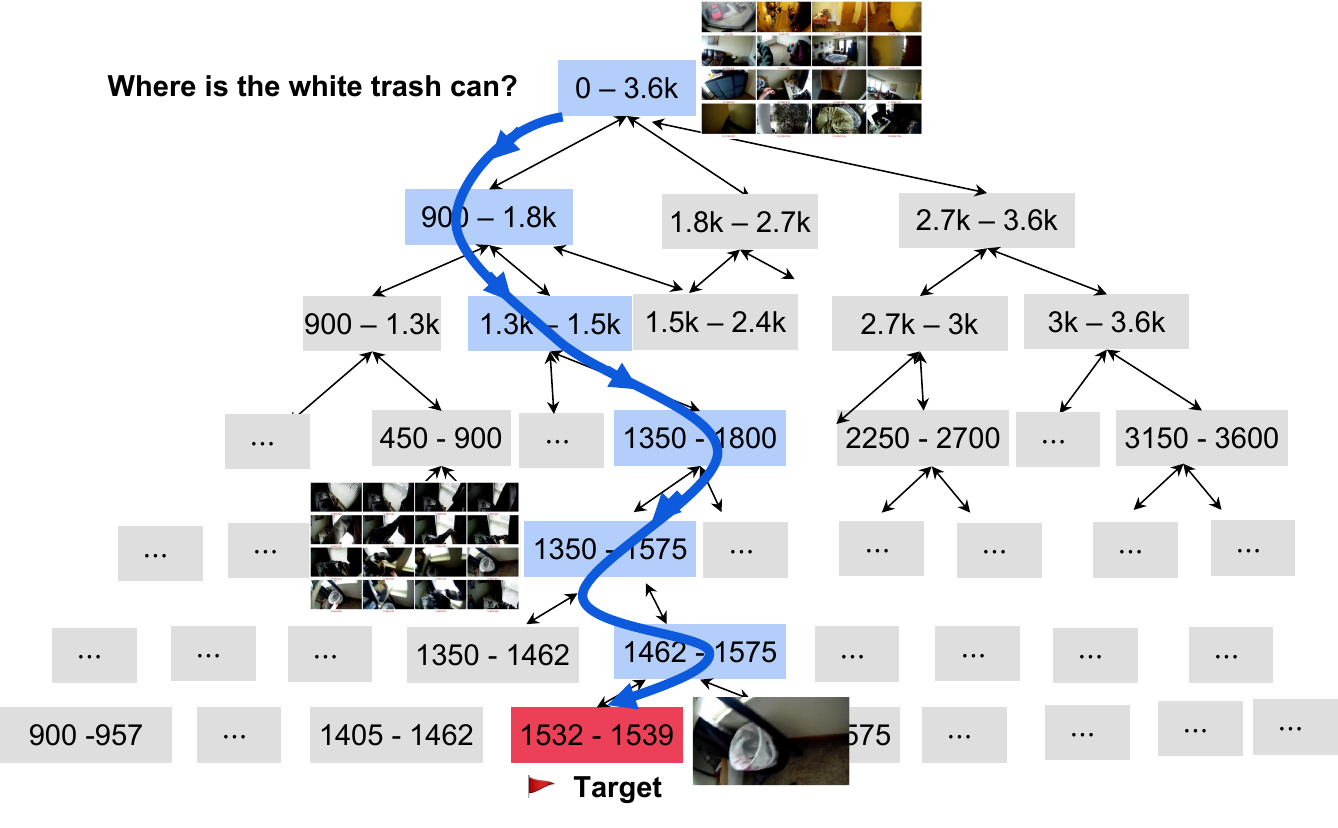}
\caption{\textbf{Illustration of the {\fancy} search process on a $b$-ary tree.} 
The video duration is 3.6k seconds, and the target white trash appears between 1532 and 1539 seconds. Numbers in the figure indicate the visited intervals of nodes, while the lines indicate the visited nodes and the search trajectory.}
\label{fig:searching_tree}
\vspace{-1em}
\end{figure}

As illustrated in Figure~\ref{fig:searching_tree}, {\fancy} is a quaternary search algorithm operating on a $b$-ary search tree. At each step, video frames are sampled on a grid with $b = n \times n$ cells. The top 25\% of regions based on their scores are retained, and the algorithm prioritizes sampling frames around these high-scoring regions. Similar to the A* algorithm, {\fancy} uses a heuristic scoring function to select branches, thereby shortening the search path. Ultimately, it performs a quaternary search with a heuristic function on a $b$-ary tree.

To simplify the discussion, assume a video of length $L$, containing only one frame that satisfies the target condition $f_t$. The grid size is $b = 2 \times 2$, and the probability that the scoring function selects the correct branch is $P$. The complexity analysis is conducted for the worst-case, best-case, and average-case scenarios.

\noindent
\textbf{Worst Case ($P \leq \frac{1}{b}$)}:
In the worst case, when $P \leq \frac{1}{b}$, the scoring function provides no effective guidance, effectively selecting branches at random. The algorithm degrades to a linear search, sequentially checking each frame until the target frame $f_t$ is found. The time complexity can be expressed as:
\begin{equation}
    T_{\text{worst}} = \mathcal{O}(L),
\end{equation}
where $L$ is the total number of video frames.

\noindent
\textbf{Best Case ($P = 1$)}:
In the best case, when $P = 1$, the scoring function always selects the correct branch leading towards the target frame. The algorithm approaches the target frame directly at each step, similar to a $b$-ary search. The time complexity is given by:
\begin{equation}
    T_{\text{best}} = \mathcal{O}(\log_b L),
\end{equation}
where $b = n \times n$ is the branching factor determined by the grid size.

\noindent
\textbf{General  Case ($\frac{1}{b} < P < 1$)}:
In the general case, the scoring function improves branch selection accuracy based on scene correlations (e.g., a kitchen scene is more likely to contain a refrigerator than a bed). The search process can be modeled as a tree with depth:
\begin{equation}
    m = \log_b L,
\end{equation}
where $m$ represents the depth of the tree. At each level, the expected number of attempts to correctly select the branch is $\frac{1}{P}$. Therefore, the total expected number of nodes visited is:
\begin{equation}
    E[N] = m \times \frac{1}{P},
\end{equation}
where $E[N]$ represents the expected number of nodes visited.

The average time complexity is then given by:
\begin{equation}
    T_{\text{avg}} = \mathcal{O}\left( \frac{\log_b L}{P} \right),
\end{equation}
where the efficiency of the algorithm is inversely proportional to the scoring function’s accuracy $P$. A higher $P$ value reduces the exploration of incorrect branches, significantly improving efficiency.

\begin{figure}
    \centering
    \includegraphics[width=0.95\linewidth]{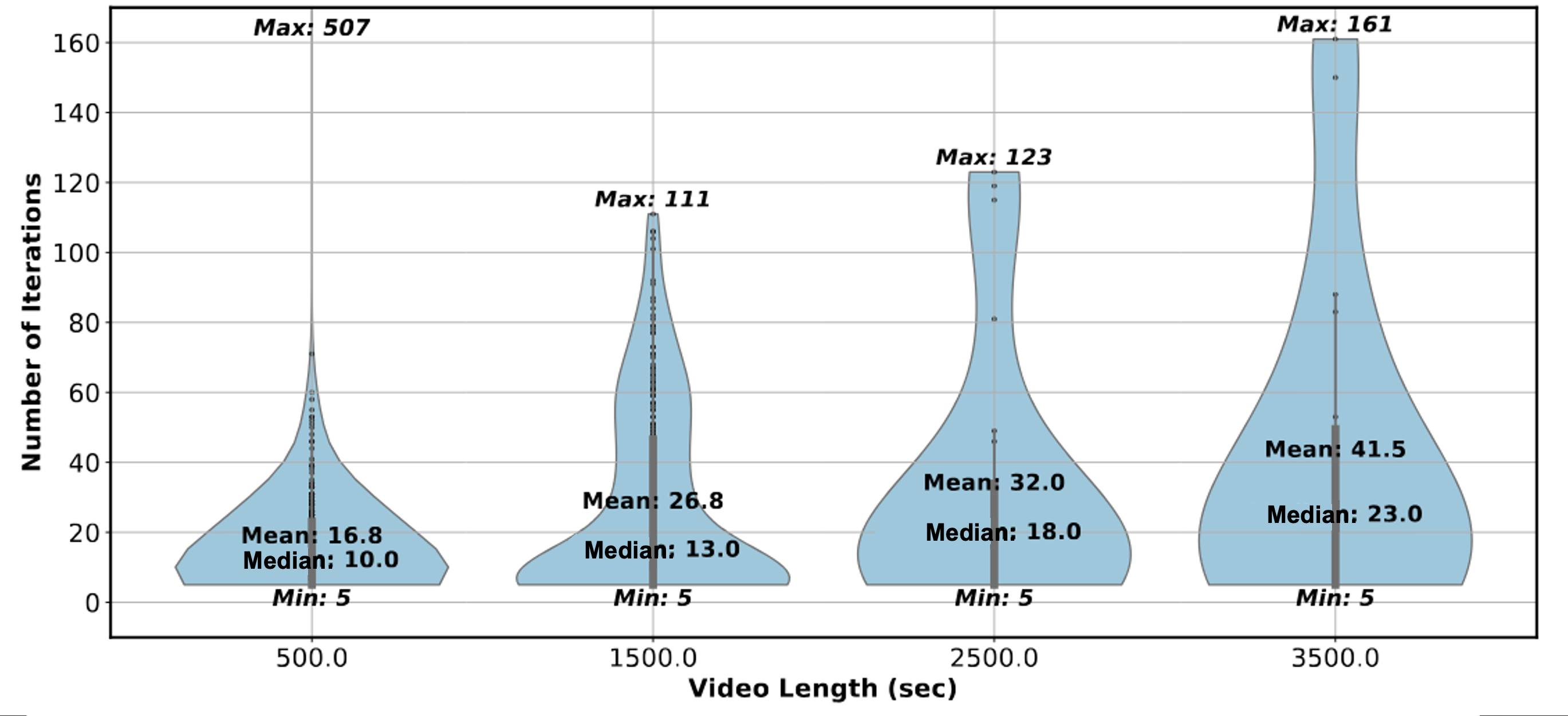}
    \caption{\textbf{Empirical results on search complexity of {\fancy}}. We show the number of iterations of $T^*$ on the LongVideoBench dataset, dividing videos with lengths between 0 and 4000 seconds into four groups. The figure shows the generally required intensity of iterative search as the length of videos vary.  
    }
        \label{fig:enter-label}
\end{figure}

Figure~\ref{fig:enter-label} shows the behavior of \fancy across various video lengths, presenting empirical statistics of search steps from our \bench dataset. The statistical results demonstrate that for videos ranging from 100 to 3600 seconds, the average number of search steps is categorized into four equidistant groups. The average number of steps required by \fancy increases gradually with video length. Notably, for videos longer than 3000 seconds, the maximum number of search steps recorded is 161, the minimum is 5 steps, and the average is 41.5 steps to complete the search. These variations are attributable to the differing intrinsic correlations within the content of each video, which informs the heuristic-based object detection process.

%% file: Sections/A9-3-Detail-Analysis.tex
\section{Detail Analysis on LongVideoBench} \label{sec:app_c_analysis}

\begin{table*}[ht]
    \centering
    \setlength\tabcolsep{17pt} 
    \setlength\extrarowheight{3pt} 
    \arrayrulecolor[gray]{0.7} 
    \begin{adjustbox}{width=0.98\linewidth}
    \begin{tabular}{l|c|c|c|c|c}
        \toprule
        \multicolumn{6}{c}{\bf LongVideoBench} \\
        \cmidrule{1-6}
        
        \multirow{3}{*}{\bf Model and Size} & \multirow{3}{*}{\bf \#Frame} & \multicolumn{4}{c}{\textbf{Video Length}} \\
        & & \textbf{XLong} & \textbf{Long} & \textbf{Medium} & \textbf{Short} \\
        & & 15-60min & 2-10min & 15-60s & 8-15s \\

        \midrule
        
        GPT4o & 8 & 47.1 & 49.4 & 67.3 & 69.7 \\
        \rowcolor{gray!10}
        GPT4o + {\fancy} & 8 & \textbf{51.6  $\pm$  1.4} & \textbf{51.7  $\pm$  1.7} & \textbf{72.9 $\pm$ 1.2} & \textbf{70.2 $\pm$ 0.2} \\
        \arrayrulecolor{gray!40}\hline \arrayrulecolor{black}

        LLaVA-OneVision-72B & 8 & 53.7 & 57.4 & 74.1 & 73.0 \\
        \rowcolor{gray!10}
        LLaVA-OneVision-72B + {\fancy} & 8 & \textbf{55.3 $\pm$ 1.3} & \textbf{63.5 $\pm$ 1.2} & \textbf{76.6 $\pm$ 1.3} & \textbf{73.7 $\pm$ 0.2} \\

        \midrule
        GPT4o & 32 & 50.5 & 57.3 & 73.5 & 71.4 \\
        \rowcolor{gray!10}
        GPT4o + {\fancy} & 32 & \textbf{53.3 $\pm$ 1.2} & \textbf{59.2 $\pm$ 1.2} & \textbf{74.3 $\pm$ 0.0} & \textbf{71.4 $\pm$ 0.0} \\
        \arrayrulecolor{gray!40}\hline \arrayrulecolor{black}

        LLaVA-OneVision-72B & 32 & 56.5 & 61.6 & 77.4 & 74.3 \\
        \rowcolor{gray!10}
        LLaVA-OneVision-72B + {\fancy} & 32 & \textbf{62.6 $\pm$ 1.2} & \textbf{63.9 $\pm$ 1.2} & \textbf{79.3 $\pm$ 0.0} & \textbf{74.6 $\pm$ 0.0} \\

        \bottomrule
    \end{tabular}
    \end{adjustbox}
    \caption{
    Detailed downstream task evaluation results for $T^*$ as an additional frame selection module for VLMs on LongVideoBench. The metric is QA accuracy (\%). We run {\fancy} two times and report the average accuracy and standard deviation ( $\pm$ ).
    }
    \label{tab:longvideobench_results}    
\end{table*}

Table \ref{tab:longvideobench_results} highlights the impact of incorporating {\fancy} as a frame selection module on QA accuracy across different video lengths in the LongVideoBench dataset.

\noindent
\textbf{Overall Effectiveness of {\fancy}:} Incorporating {\fancy} consistently improves QA accuracy for both GPT4o and LLaVA-OneVision-72B across all video lengths, demonstrating the effectiveness of the keyframe selection module in enhancing video understanding. For instance, in XLong videos (15-60 minutes), GPT4o's accuracy improves from 47.1 to 51.55 $\pm$ 0.35, while LLaVA-OneVision-72B's accuracy increases from 53.7 to 55.25 $\pm$ 0.25.

\noindent
\textbf{Impact of Video Length:} The improvements are more pronounced for longer videos (XLong and Long), where information density is higher, suggesting that {\fancy} is particularly effective in identifying and prioritizing relevant frames in complex scenarios. For shorter videos (Medium and Short), while the improvements are relatively smaller, {\fancy} still contributes to stabilizing performance across multiple runs.

\noindent
\textbf{Effect of Model Size:} Larger models, such as LLaVA-OneVision-72B, benefit slightly more from {\fancy} compared to smaller models like GPT4o, especially for longer videos. This indicates that larger models can better utilize the high-quality keyframes selected by {\fancy}.

In conclusion, {\fancy} consistently enhances QA accuracy across various video lengths, with greater impact on longer videos and larger models. These results demonstrate the potential of {\fancy} in improving video-language understanding and reasoning in long-form videos.

%% file: Sections/A9-4-Implementation-Details.tex
\section{Implementation Details}
\label{sec:Implementation}

\subsection{Implementation of Training-free \textbf{{\fancy}}}

\paragraph{Question Grounding:} 
For Question Grounding, we primarily use the \llava-OneVision 7B model, applying it to 8 uniformly sampled frames. The prompt adheres to the official release guidelines, and the specific template used is listed in Table~\ref{tab:prompt_query_grounding}.

\noindent
\paragraph{Iterative Temporal Search:} 
The default configuration for the image grid size is \(b = 8 \times 8\). We set the return threshold \(\theta\) at 0.6 for object-based and training-based scoring functions as trade-off in Figure~\ref{fig:threshold_impact}. For the attention-based method, we typically use the sum of the attention scores from the target object in the last layer of each frame. This approach was chosen because using smaller models or shallower layers resulted in performance below the baseline. Additionally, the process terminates after three iterations to manage the high computational costs associated with using the 72B model.

\noindent
\paragraph{Downstream Question Answering:} 
For downstream task evaluations, we experiment with the most prominent state-of-the-art (SOTA) models, both open- and closed-source, namely GPT4o and \llava-OneVision 72B. For GPT4o, we use the official API. For \llava, we employ the official code. The prompt template for this testing is listed in Table~\ref{tab:prompt_qa}.

\subsection{Implementation of Trainable $T^*$}

In our framework, both object-based and attention-based {\fancy} methods score each cell within the image grid and guide zooming based on straightforward rules. The training-based {\fancy} approach, however, renders this iterative search process learnable.

To learn the search policy, we employ a reinforcement learning approach. Its action space, reward function, and loss function are described as follows:

\noindent
\textbf{Action Space}~~To implement trainable scoring, we replace YOLO's detection header with a single-layer Multilayer Perceptron (MLP). This MLP maps high-level detection features into a score, indicating the likelihood that a specific area within a frame contains the visual context necessary to answer the question. For an image grid with $b = n \times n$ cells, each cell is assigned a predicted score, represented as $\mathcal{C} \in \mathbb{R}^{n \times n}$, which serves as the action space:
\begin{equation} 
\mathcal{C}, \mathcal{B} \gets \text{ScoreFunction}(G, \mathcal{T}) = \text{MLP(YOLO}(G, \mathcal{T})\text{)}.
\label{eqn:scorefucntion}
\end{equation}

\noindent
\textbf{Reward Function}~~To evaluate the quality of selected frames, we define a reward function based on their effectiveness in answering the question. Using the predicted scores $\mathcal{C} \in \mathbb{R}^{n \times n}$, we select $K$ frames and pass them to a VLM for question answering. The reward is calculated as the difference in accuracy between selected frames and a uniform baseline:
\begin{equation}
\text{reward} = \text{VLM}(K_{\text{selected}}, Q) - \text{VLM}(K_{\text{uniform}}, Q),
\label{eq:reward}
\end{equation}
where $\text{VLM}(K_{\text{uniform}}, Q)$ represents the baseline accuracy using uniform sampled frames, and $\text{VLM}(K_{\text{selected}}, Q)$ represents the accuracy using frames sampled based on the predicted relevance scores $\mathcal{C}$.

\noindent
\textbf{Loss Function}~~To optimize the trainable scoring mechanism, we employ a reinforcement learning-inspired approach using Monte Carlo estimation. We sample $K$ frames $M$ times based on the predicted scores $\mathcal{C}$. These sampled frames are passed into a Visual Language Model (e.g., \texttt{llava-OneVision-7B}) to answer the question. The average accuracy across these attempts is used as the reward signal.
The loss function for training is defined as:
\begin{equation}
\text{loss} = \sum_{i=1}^M \left( \text{reward}_i \times \text{CrossEntropy}(\mathcal{C}, \mathcal{C}_{\text{i}}) \right),
\end{equation}
where $\mathcal{C}_{\text{i}}$ represents binary labels for the $i$-th Monte Carlo sample, with selected cells for $K$ as 1 and others as 0, and $\text{reward}_i$ is the reward for the $i$-th sample, according to Eqn.~\ref{eq:reward}.

This formulation ensures that the model is reinforced to predict scores $\mathcal{C}$ aligning with the sampling labels $\mathcal{C}_{\text{i}}$ when the reward is positive. Conversely, when the reward is negative, the model adjusts its predictions to reduce the similarity between $\mathcal{C}$ and $\mathcal{C}_{\text{i}}$, penalizing incorrect sampling patterns.

\noindent
\textbf{Training and Inference}~~The search policy is trained on existing short videos and tested on unseen long videos. During the inference phase, {\fancy} uses the output of the trained YOLO model as a heuristic score.  All training and inference operations are carried out on a cluster of 8*H800 Nvidia GPUs.

We observed that models trained on the NExT-QA dataset can also effectively identify better frames for long video tasks, such as those in LongVideoBench. This suggests that unifying video representation as an $n \times n$ grid—whether for short or long videos—enables a consistent approach. Furthermore, identifying better frames should be considered a foundational task, facilitating cross-dataset generalization.

\subsection{Implementation of the baseline VideoAgent}

VideoAgent~\cite{wang2024videoagent}, the state-of-the-art temporal search baseline, leverages LLM-based video keyframe selection to optimize VLM input. It generates captions to describe video content and incrementally aggregates relevant information for question answering. We adapt the original public code to make it runnable for long video haystack setting and benchmark. While the original VideoAgent implementation uses BLIP-Large~\cite{li2023blip} for caption generation, which is significantly larger than our YOLO-based approach~\cite{cheng2024yolo} (110M parameters), we adapted the implementation to use CLIP-1B~\cite{2103.00020} for fair comparison. Specifically, we employed \texttt{clip-vit-large-patch14} and \texttt{blip-image-captioning-large}\footnote{Available at \href{https://huggingface.co/openai/clip-vit-large-patch14}{CLIP} and \href{https://huggingface.co/Salesforce/blip-image-captioning-large}{BLIP}} for our experiments.

\subsection{Video QA Implementation}
For downstream video question answering experiments, we uniformly use \llava-OneVision 72B~\cite{li2024llava} as our QA model. This open-source VLM excels in processing multimodal inputs, integrating text, image, and video analysis. We selected this model for its ability to handle arbitrary video frames and its demonstrated superior performance across diverse VQA benchmarks.

\begin{figure*}[t]
    \centering
    \includegraphics[width=\linewidth]{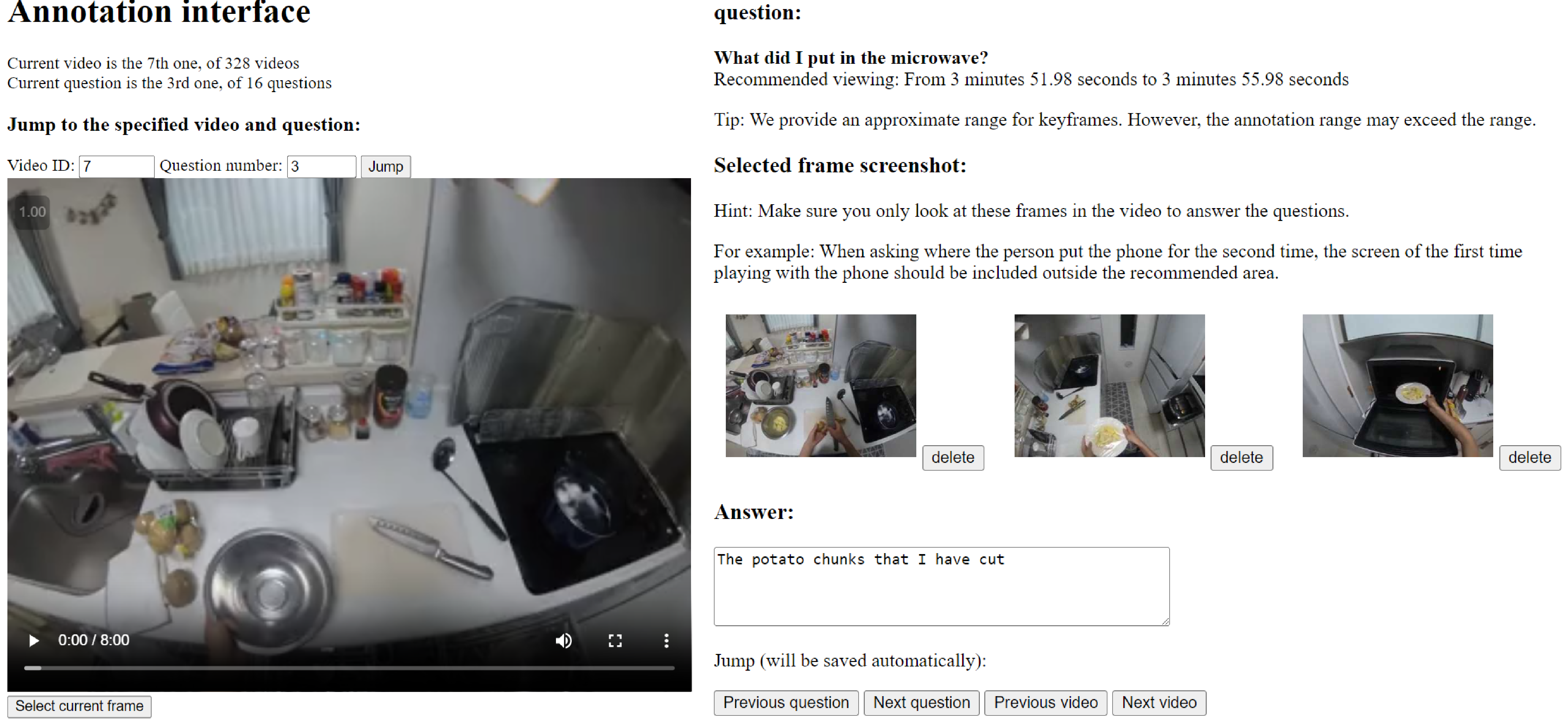}
    \caption{
    \textbf{Annotation Interface for Videos.}
        This interface allows annotators to answer questions based on video clips by keyframe annotations. Annotators can navigate to specific videos and questions using the provided controls (Video / Question ID). The current question is displayed with the recommended time range identified by the Ego4D dataset. Annotators can select key video frames and delete or modify annotations. A text box is available for entering answers based on observed video content. 
    }
    \label{fig:annotation_interface}
\end{figure*}

\subsection{Implementation of Different Search Strategies}
The core of {\fancy} leverages a well-trained open-world YOLO model for rapid object verification based on question questions. We evaluate {\fancy}'s effectiveness through three distinct search strategies:
\begin{itemize}
\item \textbf{Retrieval-based Search}: Utilizes the YOLO model~\cite{cheng2024yolo} to exhaustively scan and rank video frames based on target object detection confidence. The top 8 frames (by default) are selected as final outputs.
\item \textbf{Zooming In Search}: Implements a hierarchical approach starting with an $N \times N$ image grid matrix at low fps and resolution. The search progressively refines both fps and resolution in promising segments identified through object detection and visual cues, ultimately returning 8 frames.
\item \textbf{Trainable Search}: Adapts frame processing dynamically through YOLO model fine-tuning. Beginning with uniform sampling on an $N \times N$ image grid, it predicts correlation coefficients to guide subsequent grid sampling distributions. This process iterates three times by default, maintaining an 8-frame output. The model is trained on NExT-QA dataset and evaluated across multiple datasets.
\end{itemize}

%% file: Sections/A9-5-Annotation.tex
\section{Data Annotation Details}
\label{app:annotation_details}
To curate data for our benchmark, we repurpose established long-video understanding datasets that focused on question answering. To represent different visual scenes, we cover both egocentric and allocentric views \cite{ego4d,wu2024longvideobench}, resulting in two diverse subsets {\benchego } and {\benchlv}. This approach not only allows direct comparison with past results but also saves time and resources for extensive data curation. 
We ask crowd-source annotators to identify keyframes and answers for { \benchego}, while directly borrowing the keyframes and answers annotated from \textsc{LongVideoBench}.

\subsection{{\benchlv}}
To curate data for {\benchlv}, we utilize the frame positions from \textsc{LongVideoBench}~\cite{wu2024longvideobench} as ground-truth human-recommended frame indices. This decision is based on \textsc{LongVideoBench}'s annotation process, where annotators were required to propose questions based on given frame positions. Since \textsc{LongVideoBench} only retained frame position records in the validation set, we exclusively constructed {\benchlv} using data from the validation set. 
Furthermore, considering our focus on long-video understanding, we only included cases from the 3600-second duration group. To ensure broader applicability, we also excluded cases that referenced subtitles in their questions. 
As a result, we obtained a final set of 114 videos and 342 question pairs, none of our selected cases relied on text subtitles. You can use our script \textit{`Longvideobench2LVHaystackFormat.py'} to obtain {\benchlv} and check more detailed statistics.

\subsection{{\benchego}}
To create \benchego, we conducted data annotation on a dataset comprising 1,324 video clips, which were extracted from the original 988 videos containing a total of 15,092 questions. The video clips were pre-segmented by the Ego4D dataset to ensure that each clip contained sufficient context for answering the associated questions. This segmentation also simplified the annotation process, as shorter videos allowed annotators to better comprehend the content and efficiently identify keyframes.

The detailed instructions and interface (see Figure~\ref{fig:annotation_interface}) provided to the annotators are described in the next section, \textit{Data Annotation Interface}. Annotators were instructed to watch each video clip and answer a predefined set of questions. For every question, they were required to identify and mark several keyframes within the video that were relevant to their answers. Subsequently, they answered the questions based on these selected frames. 

To assist annotators, we provided a recommended time interval to help them quickly identify relevant frames. However, we also instructed them to watch the entire video before answering the questions, as the recommended intervals identified by the Ego4D dataset may not always be accurate. Watching the full video is crucial for ensuring logical correctness in keyframe identification. For example, some questions involve events such as \textit{"What is the second time that somebody does something?"}, requiring the annotator to identify both the first occurrence and the second occurrence to answer accurately.

In cases where a video did not provide sufficient clues to answer certain questions (potentially due to mistakes in the original dataset), annotators were instructed to respond with \textit{"Not able to answer the question"} and provide corresponding reasons (e.g., \textit{"The object does not occur in this video"}). 

Since the annotators were not native English speakers, we utilized the \texttt{googletrans} package to translate the original questions and the interface into their native language (Chinese). Similarly, their answers were translated back into English for consistency. 

To ensure the quality of the annotations, we randomly sampled 100 question-answer pairs from the annotated dataset. Only 2 obvious mistakes were identified in the sample, indicating a decent overall annotation quality. Consequently, we retained this annotation set for further analysis.

\section{Data Annotation Interface}
\label{sec:DAInterface}
Our data annotation interface facilitates annotators to provide precise answers to questions based on video clips. Key features include:
        \begin{itemize}
            \item \textbf{Video Navigation:} Annotators can jump to a specific video and question using the provided controls (e.g., Video ID and Question Number).
            \item \textbf{Question Display:} The current question is displayed prominently, along with the recommended time range for viewing relevant keyframes in the video.
            \item \textbf{Frame Selection:} Annotators can select specific video frames for reference and delete or adjust their selection as needed to support their answer.
            \item \textbf{Answer Input:} A dedicated text box allows annotators to provide their responses based on the observed content in the selected video frames.
            \item \textbf{Navigation Controls:} Quick navigation buttons enable moving between videos or questions efficiently.
        \end{itemize}
        This tool ensures accurate, contextual, and streamlined annotations for video content analysis tasks.

%% file: Sections/A9-6-Case_Statement.tex
\section{Qualitative Analysis}
\label{sec:cases}
Figure~\ref{fig:sampling_case} compares uniform sampling with T* sampling for long-format video understanding. The task involves identifying a "metal cylinder" in an hour-long video. Uniform sampling, which selects 8 frames randomly, misses key frames containing the metal cylinder.

In contrast, T* sampling focuses on semantically relevant frames, successfully capturing those featuring the metal cylinder. This highlights T* sampling's ability to prioritize critical visual information

%% file: Sections/A9-7-Prompt-Statement.tex
\section{Prompt Design}
\label{sec:prompt_sesigns}

In this section, we include the prompts designed for environment representation, focusing on question grounding and question answering tasks.

\subsection{Prompt for Question Grounding}

The following is the prompt used by our system for question grounding:

\begin{promptbox}{Prompt Template for Question Grounding}
\begin{lstlisting}[language=Python,style=mystyle]
<system prompt>
Here is a video:
<image>
<image>
<image>
...
Here is a question about the video:
Question: <Question>
Options: <Options>

When answering this question about the video:
1. What key objects to locate the answer?
   - List potential key objects (short sentences, separated by commas).
2. What cue objects might be near the key objects and might appear in the scenes?
   - List potential cue objects (short sentences, separated by commas).

Please provide your answer in two lines, directly listing the key and cue objects, separated by commas.
\end{lstlisting}
\end{promptbox}

\tcbcaption{The template of a question grounding prompt {\fancy}. \texttt{<system prompt>} is the default system instruction from {~\llava} and the \texttt{<image>} are PIL.Image objects for each frame and other text elements are strings. \label{tab:prompt_query_grounding}}

This prompt is designed to generate a representation of the environment that facilitates the grounding of queries in a structured, object-centric manner.

\subsection{Prompt for Question Answering}

The following prompt is used to answer questions based on the embodied environment representation. This design is adapted from \llava:

\begin{promptbox}{Prompt Template for Question Answering}
\begin{lstlisting}[language=Python,style=mystyle]
<system prompt>
Select the best answer to the following multiple-choice question based on the video.
<image>
<image>
<image>
...
Question:  <Question>
Options: <Options>

Answer with the option letter from the given choices directly.
\end{lstlisting}
\end{promptbox}

\tcbcaption{The template of a question answering prompt.  \label{tab:prompt_qa}}

\subsection{Prompt for Distractor Generation}

\begin{promptbox}{Prompt Template for Distractor Options}
\begin{lstlisting}[language=Python,style=mystyle]
<system prompt>
You are an expert in creating challenging multiple-choice questions. 
For the question: <question> with the correct answer <answer> generate 4 plausible but incorrect distractors that are closely related to the correct answer in context, category, or characteristics, making the question more challenging. 
Ensure that the distractors could reasonably seem correct to someone who is unsure of the answer. 
The output format should be:
"1. \"<Distractor 1>\"",
"2. \"<Distractor 2>\"",
"3. \"<Distractor 3>\"",
"4. \"<Distractor 4>\"".
\end{lstlisting}
\end{promptbox}

%% file: Sections/A9-9-Limitations-and-Broader-Impact.tex
\section*{Broad Impact}

The {\fancy} framework provides an efficient keyframe extraction solution compatible with any model or task, with applications in video summarization, healthcare training, entertainment indexing, and real-time surveillance. Its computational efficiency reduces energy consumption, aligning with sustainability goals. Additionally, the {\bench} benchmark advances standardized evaluation practices, encouraging innovation in long-form video understanding. 

Our proposed dataset is also applicable to foundation models that process entire videos. With our {\bench} dataset which consists of both training and test sets, we aim to show that keyframe supervision can act as a guiding mechanism, enabling models to first identify the most relevant keyframes from video, then use them to produce a contextually grounded answer. This approach can be more structured, efficient, and effective than directly predicting an answer from a long-form video.

%% file: main.bbl
\begin{thebibliography}{93}
\providecommand{\natexlab}[1]{#1}
\providecommand{\url}[1]{\texttt{#1}}
\expandafter\ifx\csname urlstyle\endcsname\relax
  \providecommand{\doi}[1]{doi: #1}\else
  \providecommand{\doi}{doi: \begingroup \urlstyle{rm}\Url}\fi

\bibitem[Alwassel et~al.(2021)Alwassel, Giancola, and Ghanem]{alwassel2021tsp}
Humam Alwassel, Silvio Giancola, and Bernard Ghanem.
\newblock Tsp: Temporally-sensitive pretraining of video encoders for localization tasks.
\newblock In \emph{Proceedings of the IEEE/CVF International Conference on Computer Vision}, pages 3173--3183, 2021.

\bibitem[Bai et~al.(2025)Bai, Chen, Liu, Wang, Ge, Song, Dang, Wang, Wang, Tang, Zhong, Zhu, Yang, Li, Wan, Wang, Ding, Fu, Xu, Ye, Zhang, Xie, Cheng, Zhang, Yang, Xu, and Lin]{Qwen2.5-VL}
Shuai Bai, Keqin Chen, Xuejing Liu, Jialin Wang, Wenbin Ge, Sibo Song, Kai Dang, Peng Wang, Shijie Wang, Jun Tang, Humen Zhong, Yuanzhi Zhu, Mingkun Yang, Zhaohai Li, Jianqiang Wan, Pengfei Wang, Wei Ding, Zheren Fu, Yiheng Xu, Jiabo Ye, Xi Zhang, Tianbao Xie, Zesen Cheng, Hang Zhang, Zhibo Yang, Haiyang Xu, and Junyang Lin.
\newblock Qwen2.5-vl technical report.
\newblock \emph{arXiv preprint arXiv:2502.13923}, 2025.

\bibitem[Beltagy et~al.(2020)Beltagy, Peters, and Cohan]{beltagy2020longformer}
Iz Beltagy, Matthew~E Peters, and Arman Cohan.
\newblock Longformer: The long-document transformer.
\newblock \emph{arXiv preprint arXiv:2004.05150}, 2020.

\bibitem[Briakou et~al.(2023)Briakou, Cherry, and Foster]{briakou2023searching}
Eleftheria Briakou, Colin Cherry, and George Foster.
\newblock Searching for needles in a haystack: On the role of incidental bilingualism in palm's translation capability.
\newblock \emph{arXiv preprint arXiv:2305.10266}, 2023.

\bibitem[Brunet et~al.(2011)Brunet, Vrscay, and Wang]{brunet2011mathematical}
Dominique Brunet, Edward~R Vrscay, and Zhou Wang.
\newblock On the mathematical properties of the structural similarity index.
\newblock \emph{IEEE Transactions on Image Processing}, 21\penalty0 (4):\penalty0 1488--1499, 2011.

\bibitem[Chandrasegaran et~al.(2024)Chandrasegaran, Gupta, Hadzic, Kota, He, Eyzaguirre, Durante, Li, Wu, and Fei-Fei]{hourvideo}
Keshigeyan Chandrasegaran, Agrim Gupta, Lea~M. Hadzic, Taran Kota, Jimming He, Cristóbal Eyzaguirre, Zane Durante, Manling Li, Jiajun Wu, and Li Fei-Fei.
\newblock Hourvideo: 1-hour video-language understanding, 2024.

\bibitem[Chen et~al.(2024)Chen, Wang, Chen, Zhang, Feng, Huang, Jia, and Zhu]{chen2024verified}
Houlun Chen, Xin Wang, Hong Chen, Zeyang Zhang, Wei Feng, Bin Huang, Jia Jia, and Wenwu Zhu.
\newblock Verified: A video corpus moment retrieval benchmark for fine-grained video understanding.
\newblock \emph{arXiv preprint arXiv:2410.08593}, 2024.

\bibitem[Cheng et~al.(2024)Cheng, Song, Ge, Liu, Wang, and Shan]{cheng2024yolo}
Tianheng Cheng, Lin Song, Yixiao Ge, Wenyu Liu, Xinggang Wang, and Ying Shan.
\newblock Yolo-world: Real-time open-vocabulary object detection.
\newblock In \emph{Proceedings of the IEEE/CVF Conference on Computer Vision and Pattern Recognition}, pages 16901--16911, 2024.

\bibitem[Cui et~al.(2022)Cui, Qian, Peng, Daskalaki, Chen, Guo, Sun, and Jiang]{cui2022video}
Ran Cui, Tianwen Qian, Pai Peng, Elena Daskalaki, Jingjing Chen, Xiaowei Guo, Huyang Sun, and Yu-Gang Jiang.
\newblock Video moment retrieval from text queries via single frame annotation.
\newblock In \emph{Proceedings of the 45th International ACM SIGIR Conference on Research and Development in Information Retrieval}, pages 1033--1043, 2022.

\bibitem[Dao(2024)]{dao2024flashattention}
Tri Dao.
\newblock Flashattention-2: Faster attention with better parallelism and work partitioning.
\newblock In \emph{The Twelfth International Conference on Learning Representations}, 2024.

\bibitem[Dao et~al.(2022)Dao, Fu, Ermon, Rudra, and R{\'e}]{dao2022flashattention}
Tri Dao, Dan Fu, Stefano Ermon, Atri Rudra, and Christopher R{\'e}.
\newblock Flashattention: Fast and memory-efficient exact attention with io-awareness.
\newblock \emph{Advances in Neural Information Processing Systems}, 35:\penalty0 16344--16359, 2022.

\bibitem[Di and Han(2023)]{Di2024}
Shangzhe Di and Yahong Han.
\newblock Grounded question-answering in long egocentric videos.
\newblock \emph{IEEE Transactions on Pattern Analysis and Machine Intelligence}, 45\penalty0 (11):\penalty0 14898--14912, 2023.

\bibitem[Dias et~al.(2023)Dias, Laureano, and Da~Costa]{dias2023keyframe}
Nigel Joseph~Bandeira Dias, Gustavo~Teodoro Laureano, and Ronaldo~Martins Da~Costa.
\newblock Keyframe selection for visual localization and mapping tasks: A systematic literature review.
\newblock \emph{Robotics}, 12\penalty0 (3):\penalty0 88, 2023.

\bibitem[Fan et~al.(2025)Fan, Ma, Wu, Du, Li, Gao, and Li]{fan2025videoagent}
Yue Fan, Xiaojian Ma, Rujie Wu, Yuntao Du, Jiaqi Li, Zhi Gao, and Qing Li.
\newblock Videoagent: A memory-augmented multimodal agent for video understanding.
\newblock In \emph{European Conference on Computer Vision}, pages 75--92. Springer, 2025.

\bibitem[Fu et~al.(2024)Fu, Dai, Luo, Li, Ren, Zhang, Wang, Zhou, Shen, Zhang, Chen, Li, Lin, Zhao, Li, Xu, Zheng, Chen, Ji, and Sun]{fu2024videomme}
Chaoyou Fu, Yuhan Dai, Yongdong Luo, Lei Li, Shuhuai Ren, Renrui Zhang, Zihan Wang, Chenyu Zhou, Yunhang Shen, Mengdan Zhang, Peixian Chen, Yanwei Li, Shaohui Lin, Sirui Zhao, Ke Li, Tong Xu, Xiawu Zheng, Enhong Chen, Rongrong Ji, and Xing Sun.
\newblock Video-mme: The first-ever comprehensive evaluation benchmark of multi-modal llms in video analysis, 2024.

\bibitem[Gao et~al.(2023)Gao, Zhou, Ji, Zhu, Yang, and Shou]{gao2023mist}
Difei Gao, Luowei Zhou, Lei Ji, Linchao Zhu, Yi Yang, and Mike~Zheng Shou.
\newblock Mist: Multi-modal iterative spatial-temporal transformer for long-form video question answering.
\newblock In \emph{Proceedings of the IEEE/CVF conference on computer vision and pattern recognition}, pages 14773--14783, 2023.

\bibitem[Gao and Xu(2021)]{Gao_2021_ICCV}
Junyu Gao and Changsheng Xu.
\newblock Fast video moment retrieval.
\newblock In \emph{Proceedings of the IEEE/CVF International Conference on Computer Vision (ICCV)}, pages 1523--1532, 2021.

\bibitem[Grauman et~al.(2022{\natexlab{a}})Grauman, Westbury, Byrne, Chavis, Furnari, Girdhar, Hamburger, Jiang, Liu, Liu, Martin, Nagarajan, Radosavovic, Ramakrishnan, Ryan, Sharma, Wray, Xu, Xu, Zhao, Bansal, Batra, Cartillier, Crane, Do, Doulaty, Erapalli, Feichtenhofer, Fragomeni, Fu, Gebreselasie, Gonz\'alez, Hillis, Huang, Huang, Jia, Khoo, Kol\'a\v{r}, Kottur, Kumar, Landini, Li, Li, Li, Mangalam, Modhugu, Munro, Murrell, Nishiyasu, Price, Ruiz, Ramazanova, Sari, Somasundaram, Southerland, Sugano, Tao, Vo, Wang, Wu, Yagi, Zhao, Zhu, Arbel\'aez, Crandall, Damen, Farinella, Fuegen, Ghanem, Ithapu, Jawahar, Joo, Kitani, Li, Newcombe, Oliva, Park, Rehg, Sato, Shi, Shou, Torralba, Torresani, Yan, and Malik]{ego4d}
Kristen Grauman, Andrew Westbury, Eugene Byrne, Zachary Chavis, Antonino Furnari, Rohit Girdhar, Jackson Hamburger, Hao Jiang, Miao Liu, Xingyu Liu, Miguel Martin, Tushar Nagarajan, Ilija Radosavovic, Santhosh~Kumar Ramakrishnan, Fiona Ryan, Jayant Sharma, Michael Wray, Mengmeng Xu, Eric~Zhongcong Xu, Chen Zhao, Siddhant Bansal, Dhruv Batra, Vincent Cartillier, Sean Crane, Tien Do, Morrie Doulaty, Akshay Erapalli, Christoph Feichtenhofer, Adriano Fragomeni, Qichen Fu, Abrham Gebreselasie, Cristina Gonz\'alez, James Hillis, Xuhua Huang, Yifei Huang, Wenqi Jia, Weslie Khoo, J\'achym Kol\'a\v{r}, Satwik Kottur, Anurag Kumar, Federico Landini, Chao Li, Yanghao Li, Zhenqiang Li, Karttikeya Mangalam, Raghava Modhugu, Jonathan Munro, Tullie Murrell, Takumi Nishiyasu, Will Price, Paola Ruiz, Merey Ramazanova, Leda Sari, Kiran Somasundaram, Audrey Southerland, Yusuke Sugano, Ruijie Tao, Minh Vo, Yuchen Wang, Xindi Wu, Takuma Yagi, Ziwei Zhao, Yunyi Zhu, Pablo Arbel\'aez, David Crandall, Dima Damen, Giovanni~Maria
  Farinella, Christian Fuegen, Bernard Ghanem, Vamsi~Krishna Ithapu, C.~V. Jawahar, Hanbyul Joo, Kris Kitani, Haizhou Li, Richard Newcombe, Aude Oliva, Hyun~Soo Park, James~M. Rehg, Yoichi Sato, Jianbo Shi, Mike~Zheng Shou, Antonio Torralba, Lorenzo Torresani, Mingfei Yan, and Jitendra Malik.
\newblock Ego4d: Around the world in 3,000 hours of egocentric video.
\newblock In \emph{Proceedings of the IEEE/CVF Conference on Computer Vision and Pattern Recognition (CVPR)}, pages 18995--19012, 2022{\natexlab{a}}.

\bibitem[Grauman et~al.(2022{\natexlab{b}})Grauman, Westbury, Byrne, Chavis, Furnari, Girdhar, Hamburger, Jiang, Liu, Liu, et~al.]{grauman2022ego4d}
Kristen Grauman, Andrew Westbury, Eugene Byrne, Zachary Chavis, Antonino Furnari, Rohit Girdhar, Jackson Hamburger, Hao Jiang, Miao Liu, Xingyu Liu, et~al.
\newblock Ego4d: Around the world in 3,000 hours of egocentric video.
\newblock In \emph{Proceedings of the IEEE/CVF conference on computer vision and pattern recognition}, pages 18995--19012, 2022{\natexlab{b}}.

\bibitem[Hong et~al.(2024)Hong, Wang, Ding, Yu, Lv, Wang, Cheng, Huang, Ji, Xue, Zhao, Yang, Gu, Zhang, Feng, Yin, Wang, Qi, Song, Zhang, Liu, Xu, Li, Dong, and Tang]{hong2024cogvlm2visuallanguagemodels}
Wenyi Hong, Weihan Wang, Ming Ding, Wenmeng Yu, Qingsong Lv, Yan Wang, Yean Cheng, Shiyu Huang, Junhui Ji, Zhao Xue, Lei Zhao, Zhuoyi Yang, Xiaotao Gu, Xiaohan Zhang, Guanyu Feng, Da Yin, Zihan Wang, Ji Qi, Xixuan Song, Peng Zhang, Debing Liu, Bin Xu, Juanzi Li, Yuxiao Dong, and Jie Tang.
\newblock Cogvlm2: Visual language models for image and video understanding, 2024.

\bibitem[Hwang et~al.(2024)Hwang, Hejna, Sadigh, and Bisk]{2409.10683}
Minyoung Hwang, Joey Hejna, Dorsa Sadigh, and Yonatan Bisk.
\newblock Motif: Motion instruction fine-tuning, 2024.

\bibitem[Jin et~al.(2024)Jin, Takanobu, Zhang, Cao, and Yuan]{jin2024chat}
Peng Jin, Ryuichi Takanobu, Wancai Zhang, Xiaochun Cao, and Li Yuan.
\newblock Chat-univi: Unified visual representation empowers large language models with image and video understanding.
\newblock In \emph{Proceedings of the IEEE/CVF Conference on Computer Vision and Pattern Recognition}, pages 13700--13710, 2024.

\bibitem[Kahatapitiya et~al.(2024)Kahatapitiya, Ranasinghe, Park, and Ryoo]{kahatapitiya2024language}
Kumara Kahatapitiya, Kanchana Ranasinghe, Jongwoo Park, and Michael~S Ryoo.
\newblock Language repository for long video understanding.
\newblock \emph{arXiv preprint arXiv:2403.14622}, 2024.

\bibitem[Kudo et~al.(2023)Kudo, Nagasawa, Suzuki, and Shimizu]{kudo2023challenging}
Keito Kudo, Haruki Nagasawa, Jun Suzuki, and Nobuyuki Shimizu.
\newblock A challenging multimodal video summary: Simultaneously extracting and generating keyframe-caption pairs from video.
\newblock In \emph{Proceedings of the 2023 Conference on Empirical Methods in Natural Language Processing}, pages 7380--7402, 2023.

\bibitem[Kuratov et~al.(2024)Kuratov, Bulatov, Anokhin, Sorokin, Sorokin, and Burtsev]{2402.10790}
Yuri Kuratov, Aydar Bulatov, Petr Anokhin, Dmitry Sorokin, Artyom Sorokin, and Mikhail Burtsev.
\newblock In search of needles in a 11m haystack: Recurrent memory finds what llms miss, 2024.

\bibitem[Lei et~al.(2018)Lei, Yu, Bansal, and Berg]{Lei2018}
Jie Lei, Licheng Yu, Mohit Bansal, and Tamara Berg.
\newblock Tvqa: Localized, compositional video question answering.
\newblock In \emph{Proceedings of the 2018 Conference on Empirical Methods in Natural Language Processing}, 2018.

\bibitem[Lei et~al.(2021)Lei, Berg, and Bansal]{lei2021detecting}
Jie Lei, Tamara~L Berg, and Mohit Bansal.
\newblock Detecting moments and highlights in videos via natural language queries.
\newblock \emph{Advances in Neural Information Processing Systems}, 34:\penalty0 11846--11858, 2021.

\bibitem[Li et~al.(2024{\natexlab{a}})Li, Zhang, Guo, Zhang, Li, Zhang, Zhang, Li, Liu, and Li]{li2024llava}
Bo Li, Yuanhan Zhang, Dong Guo, Renrui Zhang, Feng Li, Hao Zhang, Kaichen Zhang, Yanwei Li, Ziwei Liu, and Chunyuan Li.
\newblock Llava-onevision: Easy visual task transfer.
\newblock \emph{arXiv preprint arXiv:2408.03326}, 2024{\natexlab{a}}.

\bibitem[Li et~al.(2024{\natexlab{b}})Li, Shao, Xie, Xing, Ma, Stoica, Gonzalez, and Zhang]{li2024distflashattn}
Dacheng Li, Rulin Shao, Anze Xie, Eric~P Xing, Xuezhe Ma, Ion Stoica, Joseph~E Gonzalez, and Hao Zhang.
\newblock Distflashattn: Distributed memory-efficient attention for long-context llms training.
\newblock In \emph{First Conference on Language Modeling}, 2024{\natexlab{b}}.

\bibitem[Li et~al.(2023{\natexlab{a}})Li, Li, Savarese, and Hoi]{li2023blip}
Junnan Li, Dongxu Li, Silvio Savarese, and Steven Hoi.
\newblock Blip-2: Bootstrapping language-image pre-training with frozen image encoders and large language models.
\newblock In \emph{International conference on machine learning}, pages 19730--19742. PMLR, 2023{\natexlab{a}}.

\bibitem[Li et~al.(2023{\natexlab{b}})Li, He, Wang, Li, Wang, Luo, Wang, Wang, and Qiao]{li2023videochat}
KunChang Li, Yinan He, Yi Wang, Yizhuo Li, Wenhai Wang, Ping Luo, Yali Wang, Limin Wang, and Yu Qiao.
\newblock Videochat: Chat-centric video understanding.
\newblock \emph{arXiv preprint arXiv:2305.06355}, 2023{\natexlab{b}}.

\bibitem[Li et~al.(2024{\natexlab{c}})Li, Wang, He, Li, Wang, Liu, Wang, Xu, Chen, Luo, et~al.]{li2024mvbench}
Kunchang Li, Yali Wang, Yinan He, Yizhuo Li, Yi Wang, Yi Liu, Zun Wang, Jilan Xu, Guo Chen, Ping Luo, et~al.
\newblock Mvbench: A comprehensive multi-modal video understanding benchmark.
\newblock In \emph{Proceedings of the IEEE/CVF Conference on Computer Vision and Pattern Recognition}, pages 22195--22206, 2024{\natexlab{c}}.

\bibitem[Li et~al.(2024{\natexlab{d}})Li, Chen, Hu, and Zhang]{li2024llms}
Yunxin Li, Xinyu Chen, Baotain Hu, and Min Zhang.
\newblock Llms meet long video: Advancing long video question answering with an interactive visual adapter in llms.
\newblock \emph{arXiv preprint arXiv:2402.13546}, 2024{\natexlab{d}}.

\bibitem[Li et~al.(2024{\natexlab{e}})Li, Wen, Ryoo, et~al.]{li2024end}
Yang Li, Yuxin Wen, Michael~S. Ryoo, et~al.
\newblock End-to-end video question answering with frame scoring mechanisms and adaptive sampling.
\newblock \emph{arXiv preprint arXiv:2407.15047}, 2024{\natexlab{e}}.

\bibitem[Li et~al.(2024{\natexlab{f}})Li, Wen, Ryoo, et~al.]{li2024needle}
Yang Li, Yuxin Wen, Michael~S. Ryoo, et~al.
\newblock Needle in a video haystack: A scalable synthetic evaluator for video mllms.
\newblock \emph{arXiv preprint arXiv:2406.09367}, 2024{\natexlab{f}}.

\bibitem[Liang et~al.(2024{\natexlab{a}})Liang, Meng, Wang, Liu, Liu, and Zhao]{Liang2024}
Jianxin Liang, Xiaojun Meng, Yueqian Wang, Chang Liu, Qun Liu, and Dongyan Zhao.
\newblock End-to-end video question answering with frame scoring mechanisms and adaptive sampling.
\newblock In \emph{arXiv preprint arXiv:2407.15047}, 2024{\natexlab{a}}.

\bibitem[Liang et~al.(2024{\natexlab{b}})Liang, Meng, Wang, Liu, Liu, and Zhao]{liang2024end}
Jianxin Liang, Xiaojun Meng, Yueqian Wang, Chang Liu, Qun Liu, and Dongyan Zhao.
\newblock End-to-end video question answering with frame scoring mechanisms and adaptive sampling.
\newblock \emph{arXiv preprint arXiv:2407.15047}, 2024{\natexlab{b}}.

\bibitem[Lin et~al.(2023)Lin, Ye, Zhu, Cui, Ning, Jin, and Yuan]{lin2023video}
Bin Lin, Yang Ye, Bin Zhu, Jiaxi Cui, Munan Ning, Peng Jin, and Li Yuan.
\newblock Video-llava: Learning united visual representation by alignment before projection.
\newblock \emph{arXiv preprint arXiv:2311.10122}, 2023.

\bibitem[Liu et~al.(2024{\natexlab{a}})Liu, Feng, Wang, Wang, Liu, Zhao, Dengr, Ruan, Dai, Guo, et~al.]{liu2024deepseek}
Aixin Liu, Bei Feng, Bin Wang, Bingxuan Wang, Bo Liu, Chenggang Zhao, Chengqi Dengr, Chong Ruan, Damai Dai, Daya Guo, et~al.
\newblock Deepseek-v2: A strong, economical, and efficient mixture-of-experts language model.
\newblock \emph{arXiv preprint arXiv:2405.04434}, 2024{\natexlab{a}}.

\bibitem[Liu et~al.(2023)Liu, Zaharia, and Abbeel]{liu2023ring}
Hao Liu, Matei Zaharia, and Pieter Abbeel.
\newblock Ring attention with blockwise transformers for near-infinite context.
\newblock \emph{arXiv preprint arXiv:2310.01889}, 2023.

\bibitem[Liu et~al.(2024{\natexlab{b}})Liu, Li, Wu, and Lee]{liu2024visual}
Haotian Liu, Chunyuan Li, Qingyang Wu, and Yong~Jae Lee.
\newblock Visual instruction tuning.
\newblock \emph{Advances in neural information processing systems}, 36, 2024{\natexlab{b}}.

\bibitem[Liu et~al.(2024{\natexlab{c}})Liu, Zaharia, and Abbeel]{liu2024ringattention}
Hao Liu, Matei Zaharia, and Pieter Abbeel.
\newblock Ringattention with blockwise transformers for near-infinite context.
\newblock In \emph{The Twelfth International Conference on Learning Representations}, 2024{\natexlab{c}}.

\bibitem[Mangalam et~al.(2023)Mangalam, Akshulakov, and Malik]{mangalam2023egoschema}
Karttikeya Mangalam, Raiymbek Akshulakov, and Jitendra Malik.
\newblock Egoschema: A diagnostic benchmark for very long-form video language understanding.
\newblock In \emph{Thirty-seventh Conference on Neural Information Processing Systems Datasets and Benchmarks Track}, 2023.

\bibitem[Minderer et~al.(2022)Minderer, Gritsenko, Stone, Neumann, Weissenborn, Dosovitskiy, Mahendran, Arnab, Dehghani, Shen, et~al.]{minderer2022simple}
Matthias Minderer, Alexey Gritsenko, Austin Stone, Maxim Neumann, Dirk Weissenborn, Alexey Dosovitskiy, Aravindh Mahendran, Anurag Arnab, Mostafa Dehghani, Zhuoran Shen, et~al.
\newblock Simple open-vocabulary object detection.
\newblock In \emph{European Conference on Computer Vision}, pages 728--755. Springer, 2022.

\bibitem[Nelson et~al.(2024{\natexlab{a}})Nelson, Kollias, Das, Chaudhury, and Dan]{2407.01437}
Elliot Nelson, Georgios Kollias, Payel Das, Subhajit Chaudhury, and Soham Dan.
\newblock Needle in the haystack for memory based large language models, 2024{\natexlab{a}}.

\bibitem[Nelson et~al.(2024{\natexlab{b}})Nelson, Kollias, Das, Chaudhury, and Dan]{nelson2024needle}
Elliot Nelson, Georgios Kollias, Payel Das, Subhajit Chaudhury, and Soham Dan.
\newblock Needle in the haystack for memory based large language models.
\newblock \emph{arXiv preprint arXiv:2407.01437}, 2024{\natexlab{b}}.

\bibitem[OpenAI et~al.(2023)OpenAI, Achiam, Adler, Agarwal, Ahmad, Akkaya, Aleman, Almeida, Altenschmidt, Altman, Anadkat, Avila, Babuschkin, Balaji, Balcom, Baltescu, Bao, Bavarian, Belgum, Bello, Berdine, Bernadett-Shapiro, Berner, Bogdonoff, Boiko, Boyd, Brakman, Brockman, Brooks, Brundage, Button, Cai, Campbell, Cann, Carey, Carlson, Carmichael, Chan, Chang, Chantzis, Chen, Chen, Chen, Chen, Chen, Chess, Cho, Chu, Chung, Cummings, Currier, Dai, Decareaux, Degry, Deutsch, Deville, Dhar, Dohan, Dowling, Dunning, Ecoffet, Eleti, Eloundou, Farhi, Fedus, Felix, Fishman, Forte, Fulford, Gao, Georges, Gibson, Goel, Gogineni, Goh, Gontijo-Lopes, Gordon, Grafstein, Gray, Greene, Gross, Gu, Guo, Hallacy, Han, Harris, He, Heaton, Heidecke, Hesse, Hickey, Hickey, Hoeschele, Houghton, Hsu, Hu, Hu, Huizinga, Jain, Jain, Jang, Jiang, Jiang, Jin, Jin, Jomoto, Jonn, Jun, Kaftan, Łukasz Kaiser, Kamali, Kanitscheider, Keskar, Khan, Kilpatrick, Kim, Kim, Kim, Kirchner, Kiros, Knight, Kokotajlo, Łukasz Kondraciuk, Kondrich,
  Konstantinidis, Kosic, Krueger, Kuo, Lampe, Lan, Lee, Leike, Leung, Levy, Li, Lim, Lin, Lin, Litwin, Lopez, Lowe, Lue, Makanju, Malfacini, Manning, Markov, Markovski, Martin, Mayer, Mayne, McGrew, McKinney, McLeavey, McMillan, McNeil, Medina, Mehta, Menick, Metz, Mishchenko, Mishkin, Monaco, Morikawa, Mossing, Mu, Murati, Murk, Mély, Nair, Nakano, Nayak, Neelakantan, Ngo, Noh, Ouyang, O'Keefe, Pachocki, Paino, Palermo, Pantuliano, Parascandolo, Parish, Parparita, Passos, Pavlov, Peng, Perelman, de~Avila Belbute~Peres, Petrov, de~Oliveira~Pinto, Michael, Pokorny, Pokrass, Pong, Powell, Power, Power, Proehl, Puri, Radford, Rae, Ramesh, Raymond, Real, Rimbach, Ross, Rotsted, Roussez, Ryder, Saltarelli, Sanders, Santurkar, Sastry, Schmidt, Schnurr, Schulman, Selsam, Sheppard, Sherbakov, Shieh, Shoker, Shyam, Sidor, Sigler, Simens, Sitkin, Slama, Sohl, Sokolowsky, Song, Staudacher, Such, Summers, Sutskever, Tang, Tezak, Thompson, Tillet, Tootoonchian, Tseng, Tuggle, Turley, Tworek, Uribe, Vallone, Vijayvergiya,
  Voss, Wainwright, Wang, Wang, Wang, Ward, Wei, Weinmann, Welihinda, Welinder, Weng, Weng, Wiethoff, Willner, Winter, Wolrich, Wong, Workman, Wu, Wu, Wu, Xiao, Xu, Yoo, Yu, Yuan, Zaremba, Zellers, Zhang, Zhang, Zhao, Zheng, Zhuang, Zhuk, and Zoph]{2303.08774}
OpenAI, Josh Achiam, Steven Adler, Sandhini Agarwal, Lama Ahmad, Ilge Akkaya, Florencia~Leoni Aleman, Diogo Almeida, Janko Altenschmidt, Sam Altman, Shyamal Anadkat, Red Avila, Igor Babuschkin, Suchir Balaji, Valerie Balcom, Paul Baltescu, Haiming Bao, Mohammad Bavarian, Jeff Belgum, Irwan Bello, Jake Berdine, Gabriel Bernadett-Shapiro, Christopher Berner, Lenny Bogdonoff, Oleg Boiko, Madelaine Boyd, Anna-Luisa Brakman, Greg Brockman, Tim Brooks, Miles Brundage, Kevin Button, Trevor Cai, Rosie Campbell, Andrew Cann, Brittany Carey, Chelsea Carlson, Rory Carmichael, Brooke Chan, Che Chang, Fotis Chantzis, Derek Chen, Sully Chen, Ruby Chen, Jason Chen, Mark Chen, Ben Chess, Chester Cho, Casey Chu, Hyung~Won Chung, Dave Cummings, Jeremiah Currier, Yunxing Dai, Cory Decareaux, Thomas Degry, Noah Deutsch, Damien Deville, Arka Dhar, David Dohan, Steve Dowling, Sheila Dunning, Adrien Ecoffet, Atty Eleti, Tyna Eloundou, David Farhi, Liam Fedus, Niko Felix, Simón~Posada Fishman, Juston Forte, Isabella Fulford, Leo
  Gao, Elie Georges, Christian Gibson, Vik Goel, Tarun Gogineni, Gabriel Goh, Rapha Gontijo-Lopes, Jonathan Gordon, Morgan Grafstein, Scott Gray, Ryan Greene, Joshua Gross, Shixiang~Shane Gu, Yufei Guo, Chris Hallacy, Jesse Han, Jeff Harris, Yuchen He, Mike Heaton, Johannes Heidecke, Chris Hesse, Alan Hickey, Wade Hickey, Peter Hoeschele, Brandon Houghton, Kenny Hsu, Shengli Hu, Xin Hu, Joost Huizinga, Shantanu Jain, Shawn Jain, Joanne Jang, Angela Jiang, Roger Jiang, Haozhun Jin, Denny Jin, Shino Jomoto, Billie Jonn, Heewoo Jun, Tomer Kaftan, Łukasz Kaiser, Ali Kamali, Ingmar Kanitscheider, Nitish~Shirish Keskar, Tabarak Khan, Logan Kilpatrick, Jong~Wook Kim, Christina Kim, Yongjik Kim, Jan~Hendrik Kirchner, Jamie Kiros, Matt Knight, Daniel Kokotajlo, Łukasz Kondraciuk, Andrew Kondrich, Aris Konstantinidis, Kyle Kosic, Gretchen Krueger, Vishal Kuo, Michael Lampe, Ikai Lan, Teddy Lee, Jan Leike, Jade Leung, Daniel Levy, Chak~Ming Li, Rachel Lim, Molly Lin, Stephanie Lin, Mateusz Litwin, Theresa Lopez, Ryan
  Lowe, Patricia Lue, Anna Makanju, Kim Malfacini, Sam Manning, Todor Markov, Yaniv Markovski, Bianca Martin, Katie Mayer, Andrew Mayne, Bob McGrew, Scott~Mayer McKinney, Christine McLeavey, Paul McMillan, Jake McNeil, David Medina, Aalok Mehta, Jacob Menick, Luke Metz, Andrey Mishchenko, Pamela Mishkin, Vinnie Monaco, Evan Morikawa, Daniel Mossing, Tong Mu, Mira Murati, Oleg Murk, David Mély, Ashvin Nair, Reiichiro Nakano, Rajeev Nayak, Arvind Neelakantan, Richard Ngo, Hyeonwoo Noh, Long Ouyang, Cullen O'Keefe, Jakub Pachocki, Alex Paino, Joe Palermo, Ashley Pantuliano, Giambattista Parascandolo, Joel Parish, Emy Parparita, Alex Passos, Mikhail Pavlov, Andrew Peng, Adam Perelman, Filipe de Avila Belbute~Peres, Michael Petrov, Henrique~Ponde de Oliveira~Pinto, Michael, Pokorny, Michelle Pokrass, Vitchyr~H. Pong, Tolly Powell, Alethea Power, Boris Power, Elizabeth Proehl, Raul Puri, Alec Radford, Jack Rae, Aditya Ramesh, Cameron Raymond, Francis Real, Kendra Rimbach, Carl Ross, Bob Rotsted, Henri Roussez,
  Nick Ryder, Mario Saltarelli, Ted Sanders, Shibani Santurkar, Girish Sastry, Heather Schmidt, David Schnurr, John Schulman, Daniel Selsam, Kyla Sheppard, Toki Sherbakov, Jessica Shieh, Sarah Shoker, Pranav Shyam, Szymon Sidor, Eric Sigler, Maddie Simens, Jordan Sitkin, Katarina Slama, Ian Sohl, Benjamin Sokolowsky, Yang Song, Natalie Staudacher, Felipe~Petroski Such, Natalie Summers, Ilya Sutskever, Jie Tang, Nikolas Tezak, Madeleine~B. Thompson, Phil Tillet, Amin Tootoonchian, Elizabeth Tseng, Preston Tuggle, Nick Turley, Jerry Tworek, Juan Felipe~Cerón Uribe, Andrea Vallone, Arun Vijayvergiya, Chelsea Voss, Carroll Wainwright, Justin~Jay Wang, Alvin Wang, Ben Wang, Jonathan Ward, Jason Wei, CJ Weinmann, Akila Welihinda, Peter Welinder, Jiayi Weng, Lilian Weng, Matt Wiethoff, Dave Willner, Clemens Winter, Samuel Wolrich, Hannah Wong, Lauren Workman, Sherwin Wu, Jeff Wu, Michael Wu, Kai Xiao, Tao Xu, Sarah Yoo, Kevin Yu, Qiming Yuan, Wojciech Zaremba, Rowan Zellers, Chong Zhang, Marvin Zhang, Shengjia
  Zhao, Tianhao Zheng, Juntang Zhuang, William Zhuk, and Barret Zoph.
\newblock Gpt-4 technical report, 2023.

\bibitem[Pang et~al.(2021)Pang, Parrish, Joshi, Nangia, Phang, Chen, Padmakumar, Ma, Thompson, He, et~al.]{pang2021quality}
Richard~Yuanzhe Pang, Alicia Parrish, Nitish Joshi, Nikita Nangia, Jason Phang, Angelica Chen, Vishakh Padmakumar, Johnny Ma, Jana Thompson, He He, et~al.
\newblock Quality: Question answering with long input texts, yes!
\newblock \emph{arXiv preprint arXiv:2112.08608}, 2021.

\bibitem[Park et~al.(2024)Park, Ranasinghe, Kahatapitiya, Ryoo, Kim, and Ryoo]{park2024too}
Jongwoo Park, Kanchana Ranasinghe, Kumara Kahatapitiya, Wonjeong Ryoo, Donghyun Kim, and Michael~S Ryoo.
\newblock Too many frames, not all useful: Efficient strategies for long-form video qa.
\newblock \emph{arXiv preprint arXiv:2406.09396}, 2024.

\bibitem[Peng et~al.(2023)Peng, Quesnelle, Fan, and Shippole]{peng2023yarn}
Bowen Peng, Jeffrey Quesnelle, Honglu Fan, and Enrico Shippole.
\newblock Yarn: Efficient context window extension of large language models.
\newblock \emph{arXiv preprint arXiv:2309.00071}, 2023.

\bibitem[Radford et~al.(2021)Radford, Kim, Hallacy, Ramesh, Goh, Agarwal, Sastry, Askell, Mishkin, Clark, Krueger, and Sutskever]{2103.00020}
Alec Radford, Jong~Wook Kim, Chris Hallacy, Aditya Ramesh, Gabriel Goh, Sandhini Agarwal, Girish Sastry, Amanda Askell, Pamela Mishkin, Jack Clark, Gretchen Krueger, and Ilya Sutskever.
\newblock Learning transferable visual models from natural language supervision, 2021.

\bibitem[Ranasinghe et~al.(2024)Ranasinghe, Li, Kahatapitiya, and Ryoo]{ranasinghe2024understanding}
Kanchana Ranasinghe, Xiang Li, Kumara Kahatapitiya, and Michael~S Ryoo.
\newblock Understanding long videos in one multimodal language model pass.
\newblock \emph{arXiv preprint arXiv:2403.16998}, 2024.

\bibitem[Rawal et~al.(2024)Rawal, Saifullah, Farr{\'e}, Basri, Jacobs, Somepalli, and Goldstein]{rawal2024cinepile}
Ruchit Rawal, Khalid Saifullah, Miquel Farr{\'e}, Ronen Basri, David Jacobs, Gowthami Somepalli, and Tom Goldstein.
\newblock Cinepile: A long video question answering dataset and benchmark.
\newblock \emph{arXiv preprint arXiv:2405.08813}, 2024.

\bibitem[Ren et~al.(2024)Ren, Yao, Li, Sun, and Hou]{ren2024timechat}
Shuhuai Ren, Linli Yao, Shicheng Li, Xu Sun, and Lu Hou.
\newblock Timechat: A time-sensitive multimodal large language model for long video understanding.
\newblock In \emph{Proceedings of the IEEE/CVF Conference on Computer Vision and Pattern Recognition}, pages 14313--14323, 2024.

\bibitem[Rodriguez et~al.(2020)Rodriguez, Marrese-Taylor, Saleh, Li, and Gould]{rodriguez2020proposal}
Cristian Rodriguez, Edison Marrese-Taylor, Fatemeh~Sadat Saleh, Hongdong Li, and Stephen Gould.
\newblock Proposal-free temporal moment localization of a natural-language query in video using guided attention.
\newblock In \emph{Proceedings of the IEEE/CVF winter conference on applications of computer vision}, pages 2464--2473, 2020.

\bibitem[Romero and Solorio(2024)]{romero2024question}
David Romero and Thamar Solorio.
\newblock Question-instructed visual descriptions for zero-shot video question answering.
\newblock \emph{arXiv preprint arXiv:2402.10698}, 2024.

\bibitem[Song et~al.(2024)Song, Chai, Wang, Zhang, Zhou, Wu, Chi, Guo, Ye, Zhang, et~al.]{song2024moviechat}
Enxin Song, Wenhao Chai, Guanhong Wang, Yucheng Zhang, Haoyang Zhou, Feiyang Wu, Haozhe Chi, Xun Guo, Tian Ye, Yanting Zhang, et~al.
\newblock Moviechat: From dense token to sparse memory for long video understanding.
\newblock In \emph{Proceedings of the IEEE/CVF Conference on Computer Vision and Pattern Recognition}, pages 18221--18232, 2024.

\bibitem[Su et~al.(2021)Su, Lu, Pan, Wen, and Liu]{su2021roformer}
Jianlin Su, Yu Lu, Shengfeng Pan, Bo Wen, and Yunfeng Liu.
\newblock Roformer: Enhanced transformer with rotary position embedding, 2021.

\bibitem[Tan et~al.(2024)Tan, Sun, Hu, Wang, Deilamsalehy, Plummer, Russell, and Saenko]{tan2024koala}
Reuben Tan, Ximeng Sun, Ping Hu, Jui-hsien Wang, Hanieh Deilamsalehy, Bryan~A Plummer, Bryan Russell, and Kate Saenko.
\newblock Koala: Key frame-conditioned long video-llm.
\newblock In \emph{Proceedings of the IEEE/CVF Conference on Computer Vision and Pattern Recognition}, pages 13581--13591, 2024.

\bibitem[Tay et~al.(2020)Tay, Dehghani, Abnar, Shen, Bahri, Pham, Rao, Yang, Ruder, and Metzler]{tay2020long}
Yi Tay, Mostafa Dehghani, Samira Abnar, Yikang Shen, Dara Bahri, Philip Pham, Jinfeng Rao, Liu Yang, Sebastian Ruder, and Donald Metzler.
\newblock Long range arena: A benchmark for efficient transformers.
\newblock \emph{arXiv preprint arXiv:2011.04006}, 2020.

\bibitem[Tian et~al.(2018)Tian, Shi, Li, Duan, and Xu]{tian2018audio}
Yapeng Tian, Jing Shi, Bochen Li, Zhiyao Duan, and Chenliang Xu.
\newblock Audio-visual event localization in unconstrained videos.
\newblock In \emph{Proceedings of the European conference on computer vision (ECCV)}, pages 247--263, 2018.

\bibitem[Tworkowski et~al.(2023)Tworkowski, Staniszewski, Pacek, Wu, Michalewski, and Miłoś]{tworkowski2023focused}
Szymon Tworkowski, Konrad Staniszewski, Mikołaj Pacek, Yuhuai Wu, Henryk Michalewski, and Piotr Miłoś.
\newblock Focused transformer: Contrastive training for context scaling, 2023.

\bibitem[Wang et~al.(2024{\natexlab{a}})Wang, Shi, Tan, Qin, Wang, Zhang, Nambi, Ganu, and Wang]{wang2024multimodal}
Hengyi Wang, Haizhou Shi, Shiwei Tan, Weiyi Qin, Wenyuan Wang, Tunyu Zhang, Akshay Nambi, Tanuja Ganu, and Hao Wang.
\newblock Multimodal needle in a haystack: Benchmarking long-context capability of multimodal large language models.
\newblock \emph{arXiv preprint arXiv:2406.11230}, 2024{\natexlab{a}}.

\bibitem[Wang et~al.(2024{\natexlab{b}})Wang, Yuan, Zhang, and Sun]{wang2024tarsier}
Jiawei Wang, Liping Yuan, Yuchen Zhang, and Haomiao Sun.
\newblock Tarsier: Recipes for training and evaluating large video description models.
\newblock \emph{arXiv preprint arXiv:2407.00634}, 2024{\natexlab{b}}.

\bibitem[Wang et~al.(2024{\natexlab{c}})Wang, He, Hong, Cheng, Zhang, Qi, Huang, Xu, Dong, Ding, et~al.]{wang2024lvbench}
Weihan Wang, Zehai He, Wenyi Hong, Yean Cheng, Xiaohan Zhang, Ji Qi, Shiyu Huang, Bin Xu, Yuxiao Dong, Ming Ding, et~al.
\newblock Lvbench: An extreme long video understanding benchmark.
\newblock \emph{arXiv preprint arXiv:2406.08035}, 2024{\natexlab{c}}.

\bibitem[Wang et~al.(2024{\natexlab{d}})Wang, Zhang, Ren, Duan, Li, Liu, Hu, Chen, Zhang, Lu, et~al.]{wang2024needle}
Weiyun Wang, Shuibo Zhang, Yiming Ren, Yuchen Duan, Tiantong Li, Shuo Liu, Mengkang Hu, Zhe Chen, Kaipeng Zhang, Lewei Lu, et~al.
\newblock Needle in a multimodal haystack.
\newblock \emph{arXiv preprint arXiv:2406.07230}, 2024{\natexlab{d}}.

\bibitem[Wang et~al.(2024{\natexlab{e}})Wang, Zhang, Zohar, and Yeung-Levy]{VideoAgent}
Xiaohan Wang, Yuhui Zhang, Orr Zohar, and Serena Yeung-Levy.
\newblock Videoagent: Long-form video understanding with large language model as agent.
\newblock \emph{European Conference on Computer Vision (ECCV)}, 2024{\natexlab{e}}.

\bibitem[Wang et~al.(2024{\natexlab{f}})Wang, Zhang, Zohar, and Yeung-Levy]{wang2024videoagent}
Xiaohan Wang, Yuhui Zhang, Orr Zohar, and Serena Yeung-Levy.
\newblock Videoagent: Long-form video understanding with large language model as agent.
\newblock \emph{arXiv preprint arXiv:2403.10517}, 2024{\natexlab{f}}.

\bibitem[Wang et~al.(2022)Wang, Li, Li, He, Huang, Zhao, Zhang, Xu, Liu, Wang, et~al.]{wang2022internvideo}
Yi Wang, Kunchang Li, Yizhuo Li, Yinan He, Bingkun Huang, Zhiyu Zhao, Hongjie Zhang, Jilan Xu, Yi Liu, Zun Wang, et~al.
\newblock Internvideo: General video foundation models via generative and discriminative learning.
\newblock \emph{arXiv preprint arXiv:2212.03191}, 2022.

\bibitem[Wang et~al.(2024{\natexlab{g}})Wang, Yu, Stengel-Eskin, Yoon, Cheng, Bertasius, and Bansal]{wang2024videotree}
Ziyang Wang, Shoubin Yu, Elias Stengel-Eskin, Jaehong Yoon, Feng Cheng, Gedas Bertasius, and Mohit Bansal.
\newblock Videotree: Adaptive tree-based video representation for llm reasoning on long videos.
\newblock \emph{arXiv preprint arXiv:2405.19209}, 2024{\natexlab{g}}.

\bibitem[Wen et~al.(2024)]{wen2024too}
Yuxin Wen et~al.
\newblock Too many frames, not all useful: Efficient strategies for long-form video qa.
\newblock \emph{arXiv preprint arXiv:2406.09396}, 2024.

\bibitem[Wu et~al.(2024{\natexlab{a}})Wu, Li, Chen, and Li]{wu2024longvideobench}
Haoning Wu, Dongxu Li, Bei Chen, and Junnan Li.
\newblock Longvideobench: A benchmark for long-context interleaved video-language understanding, 2024{\natexlab{a}}.

\bibitem[Wu and Xie(2024)]{wu2024v}
Penghao Wu and Saining Xie.
\newblock V?: Guided visual search as a core mechanism in multimodal llms.
\newblock In \emph{Proceedings of the IEEE/CVF Conference on Computer Vision and Pattern Recognition}, pages 13084--13094, 2024.

\bibitem[Wu et~al.(2024{\natexlab{b}})Wu, Biamby, Quenum, Gupta, Gonzalez, Darrell, and Chan]{2407.13766}
Tsung-Han Wu, Giscard Biamby, Jerome Quenum, Ritwik Gupta, Joseph~E. Gonzalez, Trevor Darrell, and David~M. Chan.
\newblock Visual haystacks: A vision-centric needle-in-a-haystack benchmark, 2024{\natexlab{b}}.

\bibitem[Wu et~al.(2019)Wu, Xiong, Ma, Socher, and Davis]{wu2019adaframe}
Zuxuan Wu, Caiming Xiong, Chih-Yao Ma, Richard Socher, and Larry~S Davis.
\newblock Adaframe: Adaptive frame selection for fast video recognition.
\newblock In \emph{Proceedings of the IEEE/CVF Conference on Computer Vision and Pattern Recognition}, pages 1278--1287, 2019.

\bibitem[Wu et~al.(2023)]{wu2023discovering}
Zhenyu Wu et~al.
\newblock Discovering spatio-temporal rationales for video question answering.
\newblock \emph{arXiv preprint arXiv:2307.12058}, 2023.

\bibitem[Xiao et~al.(2021{\natexlab{a}})Xiao, Shang, Yao, and Chua]{nextqa}
Junbin Xiao, Xindi Shang, Angela Yao, and Tat{-}Seng Chua.
\newblock Next-qa: Next phase of question-answering to explaining temporal actions.
\newblock In \emph{{IEEE} Conference on Computer Vision and Pattern Recognition, {CVPR} 2021, virtual, June 19-25, 2021}, pages 9777--9786. Computer Vision Foundation / {IEEE}, 2021{\natexlab{a}}.

\bibitem[Xiao et~al.(2021{\natexlab{b}})Xiao, Chen, Zhang, Ji, Shao, Ye, and Xiao]{xiao2021boundary}
Shaoning Xiao, Long Chen, Songyang Zhang, Wei Ji, Jian Shao, Lu Ye, and Jun Xiao.
\newblock Boundary proposal network for two-stage natural language video localization.
\newblock In \emph{Proceedings of the AAAI Conference on Artificial Intelligence}, pages 2986--2994, 2021{\natexlab{b}}.

\bibitem[Xu et~al.(2023)Xu, Lan, Xie, Chen, and Lu]{xu2023retrieval}
Jiaqi Xu, Cuiling Lan, Wenxuan Xie, Xuejin Chen, and Yan Lu.
\newblock Retrieval-based video language model for efficient long video question answering.
\newblock \emph{arXiv preprint arXiv:2312.04931}, 2023.

\bibitem[Yan et~al.(2023)Yan, Xiong, Nagrani, Arnab, Wang, Ge, Ross, and Schmid]{yan2023unloc}
Shen Yan, Xuehan Xiong, Arsha Nagrani, Anurag Arnab, Zhonghao Wang, Weina Ge, David Ross, and Cordelia Schmid.
\newblock Unloc: A unified framework for video localization tasks.
\newblock In \emph{Proceedings of the IEEE/CVF International Conference on Computer Vision}, pages 13623--13633, 2023.

\bibitem[Ye et~al.(2023)Ye, Jiao, Wang, Tu, and Xiong]{ye2023cross}
Jinhui Ye, Wenxiang Jiao, Xing Wang, Zhaopeng Tu, and Hui Xiong.
\newblock Cross-modality data augmentation for end-to-end sign language translation.
\newblock \emph{arXiv preprint arXiv:2305.11096}, 2023.

\bibitem[Ye et~al.(2024)Ye, Wang, Jiao, Liang, and Xiong]{ye2024improving}
Jinhui Ye, Xing Wang, Wenxiang Jiao, Junwei Liang, and Hui Xiong.
\newblock Improving gloss-free sign language translation by reducing representation density.
\newblock \emph{arXiv preprint arXiv:2405.14312}, 2024.

\bibitem[Yu et~al.(2024{\natexlab{a}})Yu, Cho, Yadav, and Bansal]{yu2024self}
Shoubin Yu, Jaemin Cho, Prateek Yadav, and Mohit Bansal.
\newblock Self-chained image-language model for video localization and question answering.
\newblock \emph{Advances in Neural Information Processing Systems}, 36, 2024{\natexlab{a}}.

\bibitem[Yu et~al.(2024{\natexlab{b}})Yu, Jin, Wang, Chen, Jin, Zuo, Xu, Sun, Zhang, Wu, et~al.]{yu2024frame}
Sicheng Yu, Chengkai Jin, Huanyu Wang, Zhenghao Chen, Sheng Jin, Zhongrong Zuo, Xioalei Xu, Zhenbang Sun, Bingni Zhang, Jiawei Wu, et~al.
\newblock Frame-voyager: Learning to query frames for video large language models.
\newblock \emph{arXiv preprint arXiv:2410.03226}, 2024{\natexlab{b}}.

\bibitem[Zaheer et~al.(2020)Zaheer, Guruganesh, Dubey, Ainslie, Alberti, Ontanon, Pham, Ravula, Wang, Yang, et~al.]{zaheer2020big}
Manzil Zaheer, Guru Guruganesh, Kumar~Avinava Dubey, Joshua Ainslie, Chris Alberti, Santiago Ontanon, Philip Pham, Anirudh Ravula, Qifan Wang, Li Yang, et~al.
\newblock Big bird: Transformers for longer sequences.
\newblock \emph{Advances in neural information processing systems}, 33:\penalty0 17283--17297, 2020.

\bibitem[Zhang et~al.(2023{\natexlab{a}})Zhang, Lu, Islam, Wang, Yu, Bansal, and Bertasius]{zhang2023simple}
Ce Zhang, Taixi Lu, Md~Mohaiminul Islam, Ziyang Wang, Shoubin Yu, Mohit Bansal, and Gedas Bertasius.
\newblock A simple llm framework for long-range video question-answering.
\newblock \emph{arXiv preprint arXiv:2312.17235}, 2023{\natexlab{a}}.

\bibitem[Zhang et~al.(2020)Zhang, Sun, Jing, and Zhou]{zhang2020span}
Hao Zhang, Aixin Sun, Wei Jing, and Joey~Tianyi Zhou.
\newblock Span-based localizing network for natural language video localization.
\newblock \emph{arXiv preprint arXiv:2004.13931}, 2020.

\bibitem[Zhang et~al.(2023{\natexlab{b}})Zhang, Li, and Bing]{zhang2023video}
Hang Zhang, Xin Li, and Lidong Bing.
\newblock Video-llama: An instruction-tuned audio-visual language model for video understanding.
\newblock \emph{arXiv preprint arXiv:2306.02858}, 2023{\natexlab{b}}.

\bibitem[Zhao et~al.(2024)Zhao, Lu, Huo, Du, Yue, Guo, Wang, Chen, and Liu]{zhao2024needle}
Zijia Zhao, Haoyu Lu, Yuqi Huo, Yifan Du, Tongtian Yue, Longteng Guo, Bingning Wang, Weipeng Chen, and Jing Liu.
\newblock Needle in a video haystack: A scalable synthetic framework for benchmarking video mllms.
\newblock \emph{arXiv preprint arXiv:2406.09367}, 2024.

\bibitem[Zhou et~al.(2021)Zhou, Lin, Li, and Zheng]{Zhou_2021_CVPR}
Jiaming Zhou, Kun-Yu Lin, Haoxin Li, and Wei-Shi Zheng.
\newblock Graph-based high-order relation modeling for long-term action recognition.
\newblock In \emph{Proceedings of the IEEE/CVF Conference on Computer Vision and Pattern Recognition (CVPR)}, pages 8984--8993, 2021.

\bibitem[Zhou et~al.(2023)Zhou, Li, Lin, and Liang]{zhou2023adafocus}
Jiaming Zhou, Hanjun Li, Kun-Yu Lin, and Junwei Liang.
\newblock Adafocus: Towards end-to-end weakly supervised learning for long-video action understanding.
\newblock \emph{arXiv preprint arXiv:2311.17118}, 2023.

\bibitem[Zhou et~al.(2024{\natexlab{a}})Zhou, Lin, Qiu, and Zheng]{10210078}
Jiaming Zhou, Kun-Yu Lin, Yu-Kun Qiu, and Wei-Shi Zheng.
\newblock Twinformer: Fine-to-coarse temporal modeling for long-term action recognition.
\newblock \emph{IEEE Transactions on Multimedia}, 2024{\natexlab{a}}.

\bibitem[Zhou et~al.(2024{\natexlab{b}})Zhou, Shu, Zhao, Wu, Xiao, Yang, Xiong, Zhang, Huang, and Liu]{zhou2024mlvu}
Junjie Zhou, Yan Shu, Bo Zhao, Boya Wu, Shitao Xiao, Xi Yang, Yongping Xiong, Bo Zhang, Tiejun Huang, and Zheng Liu.
\newblock Mlvu: A comprehensive benchmark for multi-task long video understanding.
\newblock \emph{arXiv preprint arXiv:2406.04264}, 2024{\natexlab{b}}.

\end{thebibliography}
